\documentclass[10pt,twocolumn,letterpaper]{article}

\usepackage{cvpr}              
\usepackage[dvipsnames]{xcolor}

\usepackage{indentfirst}
\definecolor{cvprblue}{rgb}{0.21,0.49,0.74}
\usepackage[pagebackref,breaklinks,colorlinks,citecolor=cvprblue]{hyperref}

\usepackage{array}
\usepackage{multirow}
\usepackage{colortbl}
\usepackage{tabularx}
\usepackage{float}
\usepackage{adjustbox}
\usepackage{listings}
\usepackage{amsthm}

\def\paperID{10151} 
\def\confName{CVPR}
\def\confYear{2024}

\definecolor{somegray}{rgb}{0.5, 0.5, 0.5}
\newcommand{\darkgrayed}[1]{\textcolor{somegray}{#1}}
\makeatletter
\newcommand*\titleheader[1]{\gdef\@titleheader{#1}}
\AtBeginDocument{%
  \let\st@red@title\@title
  \def\@title{%
    \vskip-3em
    \bgroup\normalfont\large\centering\@titleheader\par\egroup
    \vskip1.5em\st@red@title}
}
\makeatother

\titleheader{\darkgrayed{This paper has been accepted for publication at the\\
IEEE Conference on Computer Vision and Pattern Recognition (CVPR), Seattle, 2024.
\copyright IEEE}}

\title{State Space Models for Event Cameras}
\author{Nikola Zubi\'{c}, Mathias Gehrig, Davide Scaramuzza \\
Robotics and Perception Group, University of Zurich, Switzerland\\ \\
}

\begin{document}

\twocolumn[{%
\renewcommand\twocolumn[1][]{#1}%
\maketitle
\vspace{-8.5ex}
\begin{center}
    \centering
    \captionsetup{type=figure}
    
    \begin{minipage}{0.38\textwidth}
        \includegraphics[width=\linewidth]{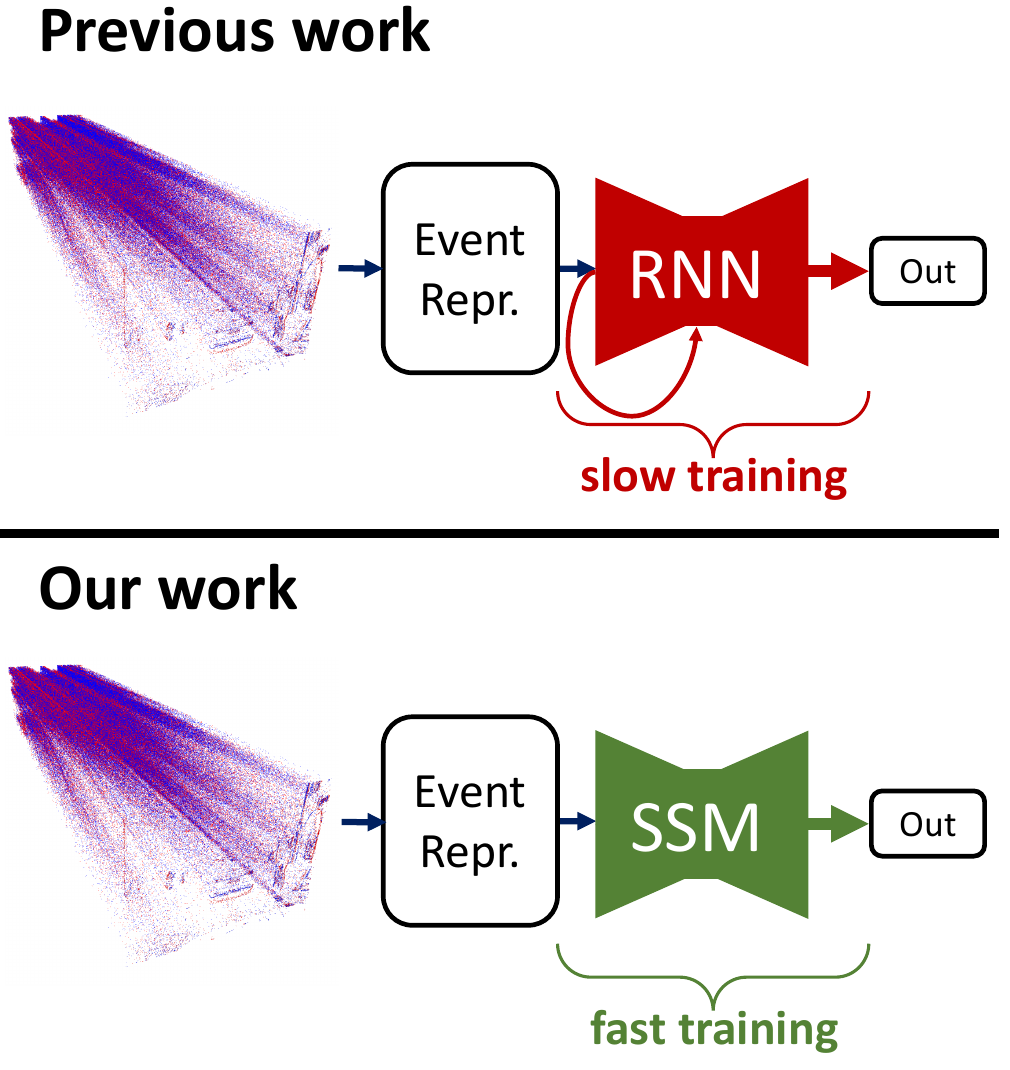}
    \end{minipage}
    \hfill 
    \begin{minipage}{0.58\textwidth}
        \includegraphics[width=\linewidth]{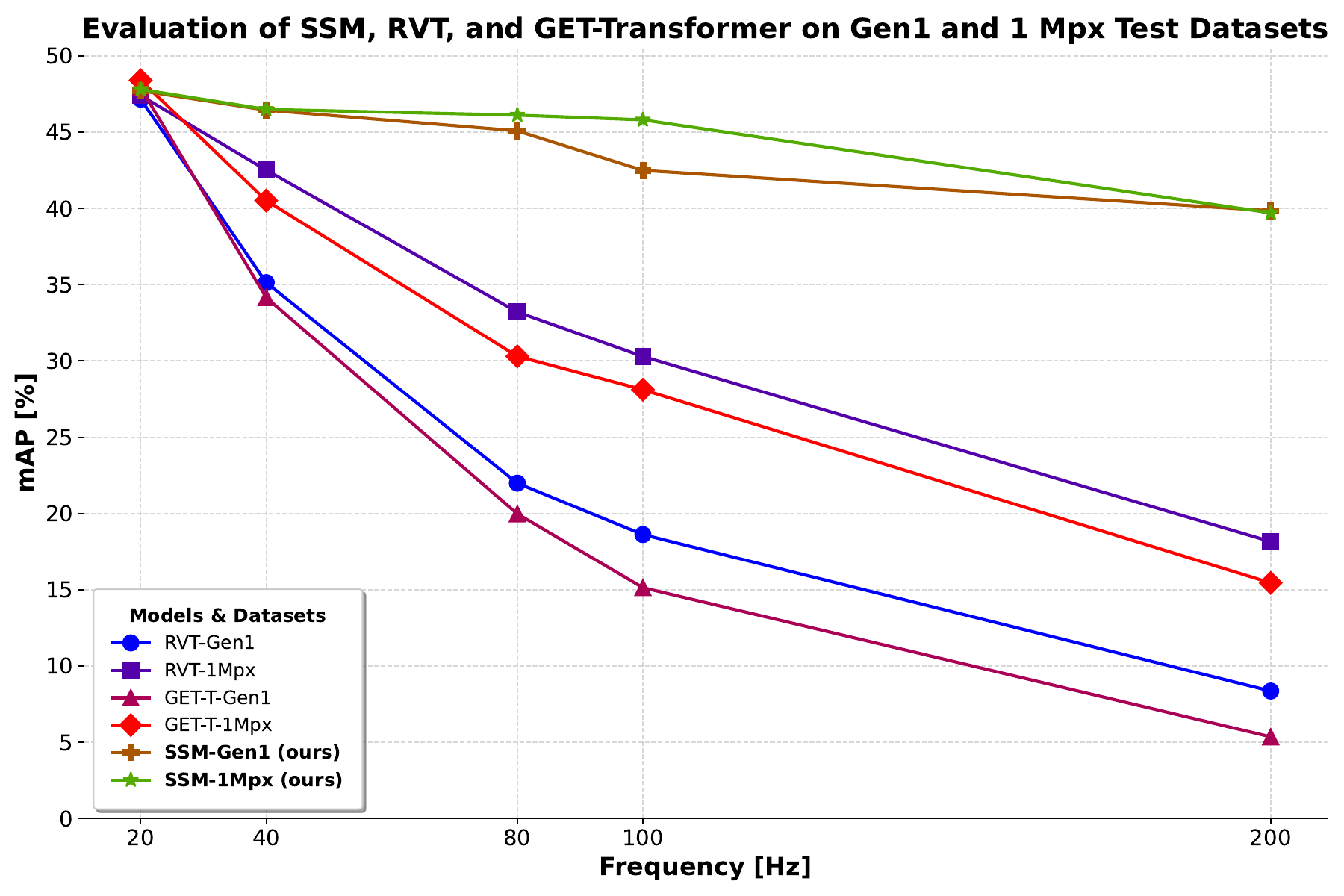}
    \end{minipage}

    \captionof{figure}{\textbf{Top-Left} Previous works \cite{Gehrig_2023_CVPR, perot_nips2020} use RNN architectures with convolutional or attention mechanisms to train models that have superior performance on downstream tasks. However, the use of RNNs leads to slower training, and the learned weights only generalize well to data deployed at the same frequency as that used at training time.
    \textbf{Bottom-Left} We solve this problem by utilizing SSMs for temporal aggregation, which enables faster training by either utilizing the S4 model \cite{gu2022efficiently} or S5 \cite{smith2023simplified} parallel scans. By their nature, these models allow deployment at different frequencies than those used at training time since they have a learnable timescale parameter. \textbf{Right} Our SSM-based models achieve an average performance drop between training and testing frequencies of 3.76 mAP averaged on both Gen1 \cite{Tournemire20arxiv} and 1 Mpx \cite{perot_nips2020} datasets, while RVT \cite{Gehrig_2023_CVPR} and GET \cite{Peng_2023_ICCV} have a drop of 21.25 and 24.53 mAP, respectively.}
\end{center}
    }]

\begin{abstract}
\vspace{-2ex}

Today, state-of-the-art deep neural networks that process event-camera data 
first convert a temporal window of events into dense, grid-like input representations. As such, they exhibit poor generalizability when deployed at higher inference frequencies (i.e., smaller temporal windows) than the ones they were trained on.
We address this challenge by introducing state-space models (SSMs) with learnable timescale parameters to event-based vision. This design adapts to varying frequencies without the need to retrain the network at different frequencies. Additionally, we investigate two strategies to counteract aliasing effects when deploying the model at higher frequencies.
We comprehensively evaluate our approach against existing methods based on RNN and Transformer architectures across various benchmarks, including Gen1 and 1 Mpx event camera datasets. Our results demonstrate that SSM-based models train 33\% faster and also exhibit minimal performance degradation when tested at higher frequencies than the training input. Traditional RNN and Transformer models exhibit performance drops of more than 20 mAP, with SSMs having a drop of 3.76 mAP, highlighting the effectiveness of SSMs in event-based vision tasks.

\end{abstract}
    
\vspace{-3ex}
\hfill \break
\noindent \textbf{Multimedial Material:} For video, code and more visit \url{https://github.com/uzh-rpg/ssms_event_cameras}. 

\vspace{-1ex}
\section{Introduction}
\label{sec:intro}
\vspace{-1ex}
Event cameras emerged as a class of sensor technologies that noticeably deviate from the operational mechanics of conventional frame-based cameras \cite{DBLP:journals/corr/abs-1904-08405}. While standard frame-based cameras \cite{zubic_aiai_2021} capture full-frame luminance levels at fixed intervals, event cameras record per-pixel relative brightness changes in a scene at the time they occur. The output is, therefore, an asynchronous stream of events in a four-dimensional spatio-temporal space.
Each event is represented by its spatial coordinates on the pixel array, the temporal instance of the occurrence, and its polarity, which denotes the direction of the brightness change and encapsulates the increase or decrease in luminance.
The power of event cameras primarily lies in their ability to capture dynamic scenes with unparalleled temporal resolution (microseconds). This property becomes invaluable in rapidly changing environments or applications requiring very fast response times \cite{Tulyakov_2021_CVPR, falanga_scirob_2020}.
However, the richness of the spatio-temporal data they generate introduces complexities in data interpretation and processing. Sophisticated algorithms are required to efficiently parse and make sense of the high-dimensional data space. As such, while event cameras represent a promising frontier in visual sensor technologies, their pervasive utilization depends upon solving these inherent computational challenges.

Current methodologies for addressing problems with event cameras fall predominantly into two categories. The first involves converting the raw stream of spatio-temporal events into dense representations akin to multi-channel images \cite{li_tpami_2023, Gehrig_2023_CVPR, Zubic_2023_ICCV, Wu2023EventCLIPAC}. This transformation allows leveraging conventional computer vision techniques designed for frame-based data. The second category employs sparse computational paradigms, often utilizing spiking neural networks or graph-based architectures \cite{Cordone_2021_IJCNN, gehrig2022pushing}. While promising, these methods are not without limitations; they frequently encounter hardware incompatibility issues and compromised accuracy. In this work, we utilize dense representations for their advantages in computational efficiency.

Despite the advances in both paradigms, models trained on event representations at a specific frequency exhibit poor generalizability when deployed in settings with higher frequencies which is crucial for high-speed, dynamical visual scenarios. Additionally, to achieve high performance, it is necessary to include recurrent memory, thereby sacrificing computational efficiency during the training phase. An ideal model would seamlessly merge the training speed of convolutional architectures with the temporal sensitivity and memory benefits inherent to recurrent models.

While recent advancements have introduced efficient recurrent vision transformer architectures \cite{Gehrig_2023_CVPR, Liu2022EventbasedMD} to achieve better performance, they face several limitations. Specifically, these architectures suffer from longer training cycles due to their reliance on conventional recurrent mechanisms.

The issue of slow training and generalization at higher event representation frequency than the lower one we trained on remains unresolved as conventional recurrent training methodologies are predominantly utilized in event-based vision. These methods do not incorporate learnable timescale parameters, thus inhibiting the model's capacity to generalize across varying inference frequencies.

In this work, we address this limitation by introducing structured variations of state-space models \cite{gu2022efficiently, smith2023simplified} as layers within our neural network framework.

State-space models \cite{gu2020hippo} function as CNN during training and are converted to an efficient RNN at test time. To achieve this, S4 \cite{gu2022efficiently} employs a technique where the SSM, which is not practical for training on modern hardware due to its sequential nature, is transformed into a CNN by unrolling. S5 \cite{smith2023simplified} uses parallel scans during training, and is employed as RNN during inference.
State-space models \cite{gu2020hippo} can be deployed at different frequencies at inference time because they are Linear Time-Invariant (LTI) continuous-time systems that can be transformed into discrete-time system with an arbitrary step size. This feature permits inference at arbitrary frequencies by globally adjusting the timescale parameter based on the ratio between the new and old frequency sampling rates. Consequently, we tackle the longstanding issue in event-based vision, which requires multiple training cycles with different frequencies to adapt the neural network for various frequencies during inference.

For the task of object detection, we find that incorporating state-space layers accelerates the training by up to 33\% relative to existing recurrent vision transformer approaches \cite{Gehrig_2023_CVPR, Liu2022EventbasedMD}. This is achieved while maintaining performance comparable to existing methods. Notably, our approach demonstrates superior generalization to higher temporal frequencies with a drop of only 3.76 mAP, while previous methods experience a drop of 21.25 mAP or more. Also, we achieve comparable performance to RVT although we use a linear state-space model rather than a non-linear LSTM model. This also shows that the complexity of the LSTM might not be needed. For this to work, we introduce two strategies (frequency-selective masking and $H_{2}$ norm) to counteract the aliasing effect encountered with increased temporal frequencies in event cameras. The first one is a low-pass bandlimiting modification to SSM that encourages the learned convolutional kernels to be smooth. The second one mitigates the aliasing problem by attenuating the frequency response after a chosen frequency. We argue that state-space modeling offers a significant new direction for research in event-based vision, offering promising solutions to the challenges inherent in processing event-based data effectively and efficiently.

Our contributions are concisely outlined as follows:
\begin{itemize}
  \setlength{\itemsep}{0pt}
  \item We introduce state-space models for event cameras to address two key challenges in event-based vision: (i) model performance degradation when operating event cameras at temporal frequencies different from their training conditions and (ii) training efficiency.
  \item Our experimental results outperform existing methods at higher frequencies by 20 mAP on average and show a 33\% increase in the training speed.
  \item We introduce two strategies (bandlimiting \& $H_{2}$ norm) designed to alleviate aliasing issues.
\end{itemize}

\section{Related Work}
\label{sec:relwork}

\subsection{Object detection with Event Cameras}

Approaches in event camera literature, thus in object detection, can be broadly classified into two branches.

The first research branch investigates dense feed-forward methods. Early attempts in this direction relied on a constrained temporal window for generating event representations \cite{Chen_2018_CVPR_Workshops, Iacono2018iros, Jiang2019icra}. The resultant models were deficient in tracking slow-moving or stationary objects, as they failed to incorporate data beyond this limited time-frame. To mitigate these drawbacks, later studies introduced recurrent neural network (RNN) layers \cite{perot_nips2020} into the architecture. The RNN component enhanced the model's capacity for temporal understanding, thereby improving its object detection capabilities. The work of Zubic et al. \cite{Zubic_2023_ICCV} takes this a step further by optimizing event representations and incorporating cutting-edge Swin transformer architecture. Nonetheless, their approach was limited in its ability to re-detect slowly moving objects following extended periods of absence. Subsequent research \cite{Gehrig_2023_CVPR, Liu2022EventbasedMD}, merged the transformer and RNN architectures, pushing the performance further. However, this significantly increased computational demands during the training phase. Importantly, all methodologies examined to date suffer from an inability to adapt when deployed at variable frequencies.

The second research branch investigates the use of Graph Neural Networks (GNNs) or Spiking Neural Networks (SNNs). GNNs dynamically construct a spatio-temporal graph where new nodes and edges are instantiated by selectively sub-sampling events and identifying pre-existing nodes that are proximate in the space-time continuum \cite{gehrig2022pushing, daobo2023, jeziorek2023}. A key challenge lies in architecting the network such that information can be disseminated effectively across extensive regions of the space-time volume. This becomes particularly important when dealing with large objects that exhibit slow relative motion to the camera. Moreover, while aggressive sub-sampling is often necessary to achieve low-latency inference, it introduces the risk of omitting important information from the event stream. On the other hand, SNNs \cite{salvatore2023, cuadrado2023, buchel2022} transmit information sparsely within the system. Unlike RNNs, SNNs emit spikes only when a threshold is met, making them hard to optimize due to the non-differentiable spike mechanism. Some solutions \cite{Messikommer20eccv} bypass the threshold, but this sacrifices sparsity in deeper layers. Overall, SNNs remain a challenging area needing more foundational research for optimized performance.

\subsection{Continuous-time Models}
Gu et al. \cite{gu2022efficiently} introduced the S4 model as an alternative to CNNs and Transformers for capturing long-range dependencies through LTI systems. This was followed by the S4D model \cite{gu_nips2022}, designed for easier understanding and implementation, offering similar performance to S4. The S5 model \cite{smith2023simplified} improved efficiency by avoiding frequency domain computations and utilizing time-domain parallel scans. However, these models have not been thoroughly evaluated on complex, high-dimensional visual data with significant temporal resolution.

In our study, we empirically show that S4, S4D, and S5 models achieve results on par with state-of-the-art when combined with attention mechanisms on complex data. We also identify and address aliasing issues in these models, proposing two corrective strategies. Our work extends the range and robustness of continuous-time models for complex sequence modeling tasks such as object detection for event cameras.
\section{Method}
\label{sec:method}
In this section, we first formalize the operating mechanism of event cameras and provide a notation used for describing state-space models in the preliminaries (Sec.~\ref{sec:method:preliminaries}). Secondly, we describe our approach of incorporating variants of state-space models as layers within our block (Sec.~\ref{sec:method:ssm_vit}). This innovative design solves the problems associated with slow training and the variable frequency inference for event cameras.
\subsection{Preliminaries}
\label{sec:method:preliminaries}
\textbf{Event cameras.}
Event cameras are bio-inspired vision sensors that capture changes in log intensity per pixel asynchronously, rather than capturing entire frames at fixed intervals. Formally, let $I(x,y,t)$ denote the log intensity at pixel coordinates $(x,y)$ and time $t$. An event $e$ is generated at $(x,y,t)$ whenever the change in log intensity $\Delta I$ exceeds a certain threshold $C$: \begin{equation}
\Delta I(x, y, t) = I(x, y, t) - I(x, y, t - \Delta t) \geq C
\end{equation}
Each event $e$ is a tuple $(x,y,t,p)$, where $(x,y)$ are the pixel coordinates, $t$ is the timestamp, and $p=\{-1, 1\}$ is the polarity of the event, indicating the direction of the intensity change.

\textbf{State-Space Models (SSMs).}
Linear State-Space Models (SSMs) form the crucial part of the backbone of our architecture, where we compare S4 \cite{gu2022efficiently}, S4D \cite{gu_nips2022}, and S5 \cite{smith2023simplified} layer variants. Given an input vector \( \mathbf{u}(t) \in \mathbb{R}^U \), a latent state vector \( \mathbf{x}(t) \in \mathbb{R}^P \), and an output vector \( \mathbf{y}(t) \in \mathbb{R}^{M} \), the governing equations of a continuous-time linear SSM can be mathematically represented as:
\begin{align}
    \frac{\mathrm{d}\mathbf{x}(t)}{\mathrm{d}t} &= \mathbf{A}\mathbf{x}(t) + \mathbf{B}\mathbf{u}(t), & \mathbf{y}(t) &= \mathbf{C}\mathbf{x}(t) + \mathbf{D}\mathbf{u}(t), \label{eq:cont_lssr}
\end{align}
The model is parameterized by a state transition matrix \( \mathbf{A} \in \mathbb{R}^{P \times P} \), an input matrix \( \mathbf{B} \in \mathbb{R}^{P \times U} \), an output matrix \( \mathbf{C} \in \mathbb{R}^{M \times P} \), and a direct transmission matrix \( \mathbf{D} \in \mathbb{R}^{M \times U} \).

Given a fixed step size \( \Delta \), this continuous-time model can be \emph{discretized} into a linear recurrence using various methods such as Euler, bilinear, or zero-order hold (ZOH):
\begin{align}
    \mathbf{x}_{k} &= \overline{\mathbf{A}}\mathbf{x}_{k-1} + \overline{\mathbf{B}}\mathbf{u}_{k}, & \mathbf{y}_{k} &= \overline{\mathbf{C}}\mathbf{x}_{k} + \overline{\mathbf{D}}\mathbf{u}_{k}, \label{eq:disc_lssr}
\end{align}
The parameters in the discrete-time model are functions of the continuous-time parameters, defined by the chosen discretization method. Details on SSMs are available in the appendix.

\begin{figure}[H]
    \centering    
    \includegraphics[width=\linewidth]{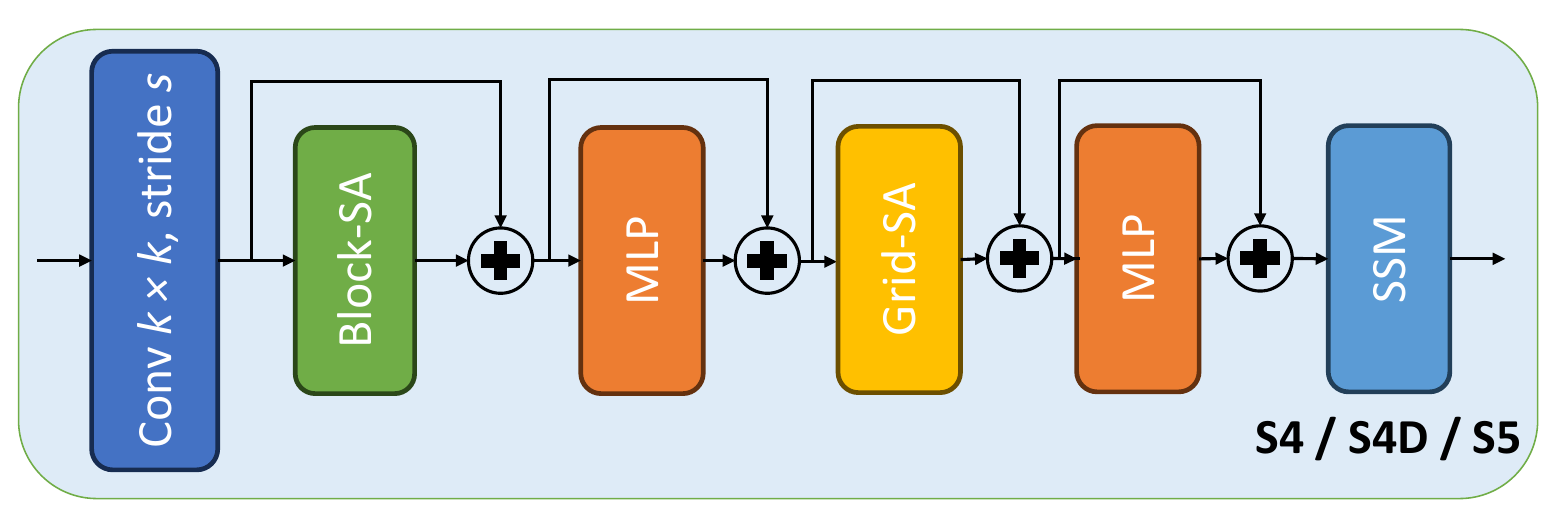}
    \caption{\textbf{SSM-ViT block structure}}
    \label{fig:ssm_vit_block}
\end{figure}

\subsection{SSM-ViT block}
\label{sec:method:ssm_vit}
In this section, we introduce the SSM-ViT block, a novel block depicted in Figure \ref{fig:ssm_vit_block}, which showcases the structured flow of the proposed block structure, designed for efficient event-based information processing. 

We build a 4-stage hierarchical backbone like in \cite{Gehrig_2023_CVPR} where in each stage we utilize our proposed SSM-ViT block. Events are processed into a tensor representation like in \cite{Gehrig_2023_CVPR} before they are used as input to the first stage. Each stage takes the previous features as input and reuses the SSM state from the last timestep to compute features for the next stage. By saving SSM states, each stage retains temporal information for the whole feature map, while also being able to generalize to different frequencies since we use SSM instead of RNN that is used in \cite{Gehrig_2023_CVPR}.

Regarding the block structure, initially, the input undergoes convolution with a defined kernel of size $k \times k$ with a stride $s$. This operation effectively captures the essential spatial features of the input. Following the convolution operation, the structure introduces a 'Block-SA' module. This module is crucial in implementing self-attention mechanisms, but it does so specifically within local windows. The localized nature of the attention in this block ensures a focus on immediate spatial relations, allowing for a detailed representation of close-proximity features.

Subsequent to 'Block-SA', the 'Grid-SA' module comes into play. In contrast to the localized approach of the previous block, 'Grid-SA' employs dilated attention, ensuring a global perspective. This module, by considering a broader scope, encapsulates a more comprehensive understanding of the spatial relations and the overall structure of the input.

The final and crucial component of the block structure is the State-Space Model (SSM) block, which is designed to compute both the output and the state in parallel, either by using S4 \cite{gu2022efficiently}, S4D \cite{gu_nips2022} or S5 \cite{smith2023simplified} model variant. This ensures temporal consistency and a seamless transition of information between consecutive time steps. Efficient computation is crucial since it allows for faster training than RNN, and the timescale parameter on temporal aggregation is important since we can rescale it during inference and deploy it at any frequency we want.

\subsection{Low-pass bandlimiting}
\label{sec:method:low_pass_bandlimiting}
However, the capability to deploy state space models at resolutions higher than those encountered during training is not without its drawbacks. It gives rise to the well-documented issue of aliasing \cite{romero2022ckconv, romero2022flexconv}, which occurs when the kernel bandwidth surpasses the Nyquist frequency. We adress this challenge in following two subsections - \ref{sec:method:low_pass_bandlimiting:masking} \& \ref{sec:method:low_pass_bandlimiting:h2_norm}.
\vspace{-1ex}
\subsubsection{Output Masking}
\label{sec:method:low_pass_bandlimiting:masking}
In the realm of signal processing, frequency content beyond the Nyquist rate can lead to aliasing, adversely affecting model performance. To address this, we integrate a frequency-selective masking strategy into the training and inference processes. This bandlimiting method has been empirically validated to be crucial for generalizing across different frequencies \cite{romero2022ckconv, romero2022flexconv}, with ablation studies indicating a decrease in accuracy by as much as 20\% in its absence.

Let the SSM be governed by a matrix $\mathbf{A}$, with its diagonal elements $a_{n}$ influencing the temporal evolution of the system states. The kernel's basis function for the $n$-th state is given by $K_n(t) = e^{t a_n} B_n$, where the frequency characteristics are primarily dictated by the imaginary part of $a_n$ - $\Im(a_{n})$ and $B_{n}$ represents the $B$ matrix.

To modulate the frequency spectrum of the model, we define a hyperparameter $\alpha$, which is used for masking the computed effective frequency $f_n$ for each state:
\begin{equation}
f_n = \frac{\Delta t}{r} \cdot \frac{|\Im(a_{n})|}{2 \pi},
\end{equation}
where $\Delta t$ denotes the discrete time-step, and $r$ is the rate at which we train the model, by default it is 1, the rate gets halved when deploying at twice the trained frequency, etc.

The bandlimiting mask is then applied as follows:
\begin{equation}
C_n = \left\{ \begin{array}{ll}
C_n & \text{if } f_n \leq \frac{\alpha}{2}, \\
0 & \text{otherwise},
\end{array} \right.
\end{equation}
where $C_n$ represents the coefficients in the state representation. The selection of $\alpha$ is critical, with $\alpha = 1.0$ representing the Nyquist limit. However, practical considerations and model constraints typically necessitate a lower empirical threshold. Our findings suggest that setting $\alpha = 0.5$ yields optimal outcomes for systems with diagonal state matrices, such as in S4D and S5 configurations.

This frequency-selective masking is proven to be a cornerstone for the adaptability of our SSMs, significantly contributing to their generalization across differing frequencies.

\vspace{-3ex}
\subsubsection{$\textbf{H}_{\textbf{2}}$ Norm}
\label{sec:method:low_pass_bandlimiting:h2_norm}
\vspace{-1ex}
This section introduces another strategy to mitigate the aliasing problem by suppressing the frequency response of the continuous-time state-space model beyond a selected frequency $\omega_\text{min}$.

This approach makes use of the $H_2$ norm of a continuous-time linear system which measures the power (or steady-state covariance) of the output response to unit white-noise input. 

Given a continuous-time system described by matrices \( \mathbf{A}, \mathbf{B}, \) and \( \mathbf{C} \), the transfer function \( \mathbf{G}(s) \) from the Laplace transform can be defined in the frequency domain as:

\begin{equation}
\mathbf{G}(j\omega) = \mathbf{C} (j\omega \mathbf{I} - \mathbf{A})^{-1} \mathbf{B},
\end{equation}

where \( \omega \) denotes the frequency, and \( \mathbf{I} \) is the identity matrix. The $H_{2}$ norm is computed as the integral of the squared magnitude of the frequency response over the range of interest, typically the entire frequency spectrum.
However, in our case, we would like to suppress the frequency response of the system to frequencies beyond $\omega_\text{min}$ which can be done by minimizing the following:

\begin{equation}
\left\| \mathbf{G} \right\|_{H_2(\omega_\text{min}, \infty)} = \sqrt{\frac{1}{\pi} \int_{\omega_\text{min}}^{\infty} \left\| \mathbf{G}(j\omega_k) \right\|_F^2 d\omega },
\end{equation}

as part of the loss function, where $\left\| \cdot \right\|_F$ is the Frobenius norm.
In practice, to numerically estimate this integral, we choose a maximum frequency $\omega_\text{max}$ and discretize the frequency range \( [\omega_{\text{min}}, \omega_{\text{max}}] \) into \( N \) points and apply numerical integration methods to the squared Frobenius norm of \( \mathbf{G}(j\omega) \). This yields an approximate $H_{2}$ norm of the system in the desired frequency range.
\vspace{-2ex}
\section{Experiments}
\label{sec:experiments}
\vspace{-1ex}
We perform ablation studies and systematic evaluations of our proposed models utilizing both the Gen1 \cite{Tournemire20arxiv} and 1 Mpx \cite{perot_nips2020} event-based camera datasets. We assess the model's proficiency in adapting to unseen inference frequencies during training time on both datasets. We employed two variants for training across both datasets: the base model ViT-SSM-B, alongside its scaled-down derivative ViT-SSM-S (small). Additionally, to study the robustness and generalization capabilities of our architecture, we subject it to empirical testing on the DSEC dataset \cite{Gehrig2021ral_a}, which was not part of the original training corpus, and the visual results are provided in the appendix.

\subsection{Setup}
\label{sec:experiments:setup}
\textbf{Implementation details.}
Our models are trained using 32-bit precision arithmetic across 400k iterations, making use of the ADAM optimizer \cite{kingma_iclr2015} and a OneCycle learning rate schedule \cite{smith_iclr2018} that decays linearly from its highest value. We adopt a mixed batching technique that applies standard Backpropagation Through Time (BPTT) to half of the batch samples and Truncated BPTT (TBPTT) to the other half. This technique was introduced in the RVT \cite{Gehrig_2023_CVPR}. We found it to be equally effective also for SSMs. For the S4(D) method, details can be found in \cite{gu2022efficiently, gu_nips2022}. As for S5 \cite{smith2023simplified}, we utilize an efficient implementation from scratch in PyTorch; the original public version of S5 is available in JAX. Data augmentation is carried out through random horizontal flips and zooming operations, both inward and outward. Event representations are formed based on 50 ms time windows that correspond to 20 Hz sampling frequency, divided into T = 10 discrete bins. We incorporate a YOLOX detection head \cite{yolox2021} that integrates IOU, class, and regression losses, averaged across both batch and sequence lengths during each optimization phase. Bandlimiting is implemented during training and inference of the SSM model by masking the output matrix \textbf{C} as explained in \ref{sec:method:low_pass_bandlimiting:masking}. In the case of \( H_2 \) norm, we add it as a term to the loss function of the model. We conduct training on the GEN1 dataset using A100 GPU, employing a batch size of 8 and a sequence length of 21. We use a global learning rate of 2e-4, which is propagated to the SSM components. For the 1 Mpx dataset, we train with a batch size of 12 and a sequence length of 10, using a learning rate of 3.5e-4 across two A100 GPUs.

\textbf{Datasets.}
\label{sec:experiments:datasets}
The Gen1 Automotive Detection dataset \cite{Tournemire20arxiv} comprises 39 hours of event camera footage with a resolution of 304 $\times$ 240 pixels. It includes 228k bounding boxes for cars and 28k for pedestrians. Using original evaluation criteria \cite{Tournemire20arxiv}, we discard bounding boxes having a side length shorter than 10 pixels or a diagonal less than 30 pixels.

The 1 Mpx dataset \cite{perot_nips2020} similarly focuses on driving environments but offers recordings at a higher resolution of 720 $\times$ 1280 pixels, captured over multiple months during both day-time and night-time. It contains roughly 15 hours of event data and accumulates around 25 million bounding box labels, spread across three classes: cars, pedestrians, and two-wheelers. Following the original evaluation criteria, we eliminate bounding boxes with a side length under 20 pixels and a diagonal less than 60 pixels, while also reducing the input resolution to 640 $\times$ 360 pixels.

\subsection{Benchmark comparisons}
\begin{table*}[ht]
    \centering
    \begin{tabular}{@{}llllllll@{}}
        \toprule
                              &                   &                & \multicolumn{2}{c|}{Gen1}           & \multicolumn{2}{c}{1 Mpx}           &            \\ \cmidrule(lr){4-7}
        Method                & Backbone          & Detection Head & mAP           & Time (ms)           & mAP           & Time (ms)           & Params (M) \\ \midrule
        Asynet \cite{Messikommer20eccv}                & Sparse CNN        & YOLOv1 \cite{Redmon_2016_CVPR}         & 14.5          & -                   & -             & -                   & 11.4       \\
        AEGNN \cite{Schaefer_2022_CVPR}                & GNN               & YOLOv1         & 16.3          & -                   & -             & -                   & 20.0       \\
        Spiking DenseNet \cite{Cordone_2022_IJCNN}      & SNN               & SSD \cite{Liu2016SSD}            & 18.9          & -                   & -             & -                   & 8.2        \\
        Inception + SSD \cite{Iacono2018iros}       & CNN               & SSD            & 30.1          & 19.4                & 34.0          & 45.2                & $>60$*      \\
        RRC-Events \cite{Chen_2018_CVPR_Workshops}            & CNN               & YOLOv3 \cite{redmon2018_arxiv}         & 30.7          & 21.5                & 34.3          & 46.4                & $> 100$*   \\
        MatrixLSTM \cite{cannici2019asynchronous}           & RNN + CNN               & YOLOv3         & 31.0          & -                   & -             & -                   & 61.5          \\
        YOLOv3 Events \cite{Jiang2019icra}         & CNN               & YOLOv3         & 31.2          & 22.3                & 34.6          & 49.4                & $>60$*      \\
        RED \cite{perot_nips2020}                   & CNN + RNN         & SSD            & 40.0          & 16.7               & 43.0          & 39.3                & 24.1       \\
        ERGO-12 \cite{Zubic_2023_ICCV}                   & Transformer         & YOLOv6 \cite{Li2022_YOLOv6}            & \textbf{50.4}          & 69.9                & 40.6          & 100.0                & 59.6       \\
        RVT-B \cite{Gehrig_2023_CVPR} & Transformer + RNN & YOLOX \cite{yolox2021}          & 47.2 & 10.2 & 47.4    & 11.9 & 18.5       \\ 
        Swin-T v2~\cite{Liu_2022_CVPR} & Transformer + RNN & YOLOX          & 45.5          & 26.6           & 46.4             & 34.5           & 21.1        \\
        Nested-T~\cite{zhang2021aggregating, Peng_2023_ICCV} & Transformer + RNN & YOLOX          & 46.3          & 25.9           & 46.0             & 33.5           & 22.2        \\
        GET-T~\cite{Peng_2023_ICCV} & Transformer + RNN & YOLOX          & \underline{47.9}          & 16.8           & \textbf{48.4}             & 18.2          & 21.9        \\
        \midrule
        \textbf{S4D-ViT-B (ours)} & Transformer + SSM & YOLOX          & 46.2          & 9.40           & 46.8             & 10.9          & 16.5        \\
        \textbf{S5-ViT-B (ours)} & Transformer + SSM & YOLOX          & 47.7          & \underline{8.16}           & \underline{47.8}            & \underline{9.57}          & 18.2       \\    \textbf{S5-ViT-S (ours)} & Transformer + SSM & YOLOX          & 46.6          & \textbf{7.81}           & 46.5           & \textbf{8.87}          & 9.7       \\      
        \bottomrule
    \end{tabular}
    \caption{\textbf{Comparisons on test sets of Gen1 and 1 Mpx datasets (20 Hz)}. Best results in \textbf{bold} and second best underlined. A star $^*$ suggests that this information was not directly available and estimated based on the publications. Runtime is measured in milliseconds for a batch size of 1. We used a T4 GPU for SSM-ViT to compare against indicated timings in prior work \cite{perot_nips2020} on comparable GPUs (Titan Xp). Our model is the 3rd best on the Gen1 and 2nd best on the 1Mpx dataset in terms of downstream task performance, while having fastest inference and less parameters.}%
    \label{tab:sota}
\end{table*}
\label{sec:experiments:benchmark}
In this section, we present a detailed comparison of our proposed methods, S4D-ViT and S5-ViT, with the SotA approaches in the domain of event-based vision for object detection, which can be seen in Table~\ref{tab:sota}. The evaluation is conducted on two distinct datasets: Gen1 and 1 Mpx datasets.

The comparative analysis, as summarized in Table~\ref{tab:sota}, encompasses different backbones and detection heads. These include Sparse CNNs, GNNs, SNNs, RNNs, and various implementations of Transformers.

Our S4D-ViT-B and S5-ViT-B models, which integrate SSMs with Attention (\ref{sec:method:ssm_vit}), demonstrate competitive performance across both datasets. Specifically, on the 1 Mpx dataset, our models register mAP scores of 46.8 for S4D and 47.8 and 46.5 for the S5 Base and Small variants, respectively. These scores are competitive with the top performances in this benchmark, securing the second-highest task performance on the 1 Mpx dataset. While our models do not outperform the top-performing ERGO-12 \cite{Zubic_2023_ICCV} and GET-T \cite{Peng_2023_ICCV} in terms of mAP on the Gen1 dataset, they are very close to GET-T \cite{Peng_2023_ICCV} with a difference of only 0.2 mAP.

Notably, our models perform consistently well across different frequencies as can be seen in Table~\ref{table:benchmark_freqs}, highlighting their robustness and generalizability in comparison to RVT and GET-T, which tend to exhibit large performance drops. We achieve an average performance drop between training and testing frequencies of 3.76 mAP averaged on both Gen1 \cite{Tournemire20arxiv} and 1 Mpx \cite{perot_nips2020} datasets, while RVT \cite{Gehrig_2023_CVPR} and GET \cite{Peng_2023_ICCV} have a drop of 21.25 and 24.53 mAP, respectively.

In summary, our proposed S4D-ViT-B and S5-ViT-B models establish themselves as effective and efficient contenders in the field of event-based vision for object detection. Their competitive performance metrics, coupled with their generalization capabilities across various inference frequencies, make them valuable contributions to the event-based vision community.
\vspace{-1ex}
\begin{table}[ht]
    \centering
    \begin{adjustbox}{max width=\linewidth}
    \setlength{\tabcolsep}{4pt}
        {\small
        \begin{tabular}{lccccccc}
        \hline
        \multirow{2}{*}{Model} & \multirow{2}{*}{Dataset} & \multicolumn{5}{c}{Frequency Evaluation (Hz)} & \multirow{2}{*}{Perf. Drop} \\
        \cline{3-7}
         &  & 20 Hz & 40 Hz & 80 Hz & 100 Hz & 200 Hz & \\
        \hline
        \multirow{2}{*}{RVT} & Gen1 & 47.16 & 35.13 & 21.98 & 18.61 & 8.35 & 26.14 \\
         & 1Mpx & 47.40 & 42.51 & 33.20 & 30.29 & 18.15 & 16.36 \\
        \hline
        \multirow{2}{*}{S5} & Gen1 & 47.71 & 46.44 & 45.08 & 42.49 & 39.84 & \underline{4.25} \\
         & 1Mpx & 47.80 & 46.49 & 46.11 & 45.80 & 39.70 & \textbf{3.27} \\
        \hline
        \multirow{2}{*}{GET} & Gen1 & 47.90 & 34.15 & 19.97 & 15.13 & 5.35 & 29.25 \\
         & 1Mpx & 48.40 & 40.51 & 30.30 & 28.11 & 15.44 & 19.81 \\
        \hline
        \end{tabular}}
    \end{adjustbox}
\caption{Evaluation of RVT \cite{Gehrig_2023_CVPR}, S5 \cite{smith2023simplified}, and GET \cite{Peng_2023_ICCV} across different frequencies on test datasets. The Performance Drop is calculated as the average difference between the original performance at 20 Hz and performances at higher frequencies.}
\label{table:benchmark_freqs}
\end{table}

\vspace{-3ex}
\subsection{Ablation study}
\label{sec:experiments:ablation}
In this section, we investigate different SSMs and their initializations with and without bandlimiting parameter for event-based vision to address the problem of inference at different frequencies (\ref{sec:experiments:ablation:ssm_inits_bandlimits}). After that, we study the importance of SSM layers in various stages during the training of the model (\ref{sec:experiments:ablation:ssm_layers}). Finally, we evaluate models on differing frequencies with two proposed strategies (\ref{sec:experiments:ablation:frequencies_comparison}).

\subsubsection{SSMs: initializations \& bandlimiting}
\label{sec:experiments:ablation:ssm_inits_bandlimits}
This section presents an ablation study focusing on the performance impact of various SSM variants and their initializations in conjunction with the bandlimiting parameter $\alpha$. The SSM variants under consideration are S4 \cite{gu2022efficiently}, S4D \cite{gu_nips2022}, and the more recent S5 \cite{smith2023simplified}. The analysis is conducted on the Gen1 \cite{Tournemire20arxiv} validation dataset, with the mAP serving as the performance metric.
Table \ref{table:bandlimit_performance_comparison} provides a comprehensive view of how different initialization strategies (legS, inv, lin) introduced in \cite{gu_nips2022} and values of $\alpha \in \{0, 0.5, 1\}$ influence model performance. The S4-legS model, not equipped for bandlimiting in non-diagonal matrix scenarios, achieves a baseline mAP of 46.66 at $\alpha=0$.
The S4D variants demonstrate a diverse performance spectrum. The S4D-legS variant, particularly at $\alpha=0.5$, achieves the highest mAP of 47.33 among the S4D models, also maintaining the best average mAP of 46.92. The S4D-inv and S4D-lin variants show less favorable outcomes, with mAPs peaking at 46.23 and 46.02, respectively, for $\alpha=0.5$.

Notably, the S5 model variants exhibit an improvement over their predecessors. The S5-legS variant achieves the highest mAP of 48.48 at $\alpha=0.5$ and also records the best average mAP of 48.27. This result is not only the best in its category but also the best overall. The S5-inv and S5-lin models also demonstrate comparable performance, with the former reaching its peak mAP of 47.43 at $\alpha=0.5$.

This study emphasizes the critical role of initialization strategies and bandlimiting in optimizing SSM-based neural networks for event-based vision tasks. The distinct performance variations across different models and configurations underscore the importance of selecting appropriate initializations and $\alpha$ values, as these choices impact the efficacy of the models in handling the dynamic and complex nature of event-based vision data. Higher values of parameter $\alpha$ encourage SSM to learn smoother kernels and discard more complex and higher-frequency convolution kernels.

\begin{table}[t]
\centering
\begin{tabular}{lcccc}
\hline
\multirow{2}{*}{Model} & \multicolumn{4}{c}{Mean Average Precision - mAP$_{val}$} \\
\cline{2-5}
 & $\alpha=0$ & $\alpha=0.5$ & $\alpha=1$ & Average \\
\hline
S4-legS & 46.66 & - & - & 46.66 \\
\hline
S4D-legS & 46.93 & 47.33 & 46.50 & 46.92 \\
S4D-inv & 46.15 & 46.23 & 46.11 & 46.16 \\
S4D-lin & 44.82 & 46.02 & 45.04 & 45.29 \\
\hline
S5-legS & 48.33 & \textbf{48.48} & 48.00 & \textbf{48.27} \\
S5-inv & 47.26 & \underline{47.43} & 46.98 & \underline{47.22} \\
S5-lin & 46.12 & 46.40 & 45.59 & 46.04 \\
\hline
\end{tabular}
\caption{Performance comparison between the S4 \cite{gu2022efficiently}, S4D \cite{gu_nips2022} and S5 \cite{smith2023simplified} models for different values of $\alpha$ and initializations on Gen1 \cite{Tournemire20arxiv} validation dataset.}
\label{table:bandlimit_performance_comparison}
\end{table}

\vspace{-1ex}
\subsubsection{SSM Utilization Analysis}
\label{sec:experiments:ablation:ssm_layers}
In this ablation study, we examine the impact of employing temporal recurrence exclusively in a subset of network stages, or not using it at all. Our methodology involves manipulating the SSMs within the network by resetting their states at predetermined stages during each timestep. This approach enables us to isolate and evaluate the impact of SSM layers' presence or absence while maintaining a consistent parameter count across different model configurations.
\begin{table}[ht]
\centering
\begin{tabularx}{\linewidth}{cccc|ccc}
\hline
S1 & S2 & S3 & S4 & mAP$_{RVT}$ & mAP$_{S4D_{.5}}$ & mAP$_{S5_{.5}}$ \\
\hline
& & & & 33.90 & 39.99 & 43.67 \\
& & & \checkmark & 41.68 & 43.11 & 46.10 \\
& & \checkmark & \checkmark & 46.10 & 45.33 & 47.52  \\
& \checkmark & \checkmark & \checkmark & 48.82 & 47.02 & 48.41 \\
\checkmark & \checkmark & \checkmark & \checkmark & 49.52 & 47.33 & 48.48 \\
\hline
\end{tabularx}
\caption{SSM contribution in various stages on the Gen1 dataset.}
\label{table:performance_comparison}
\end{table}

Table~\ref{table:performance_comparison} shows the outcomes of these manipulations on the Gen1 validation dataset. $S4D_{.5}$ represents S4D model, and $S5_{.5}$ is S5 model with $\alpha=0.5$. The data clearly indicate that the complete removal of SSMs from the network leads to the highest decrease in detection performance. This underscores the critical role of SSMs in enhancing the model's capability. On the other hand, initiating the use of SSMs from the fourth stage onwards consistently enhances performance, suggesting a critical threshold for the impact of temporal information processing in the later stages of the network.
Another intriguing observation is the performance boost obtained by incorporating an SSM at the very initial stage. This suggests that the early integration of temporal information is also beneficial for the overall detection performance. As a result of these findings, our preferred configuration includes the SSM component right from the initial stage, thereby leveraging the advantages of temporal information processing throughout the entire network. Noteworthy is the observation that the performance drop when employing SSMs is less pronounced than with RVT, suggesting our approach's superior robustness to temporal aggregation from certain stages compared to RVT.
\vspace{-2ex}
\subsubsection{Evaluation at different frequencies}
\vspace{-1.5ex}
In this section, we compare the performance of three models: RVT \cite{Gehrig_2023_CVPR}, S4D \cite{gu_nips2022}, and S5 \cite{smith2023simplified}, across various frequency ranges. This evaluation is crucial in determining the robustness and adaptability of these models under diverse inference frequencies, specifically at 20 Hz, 40 Hz, 80 Hz, 100 Hz, and 200 Hz. Our model is trained with event representations formed based on 50 ms time windows that correspond to a 20 Hz sampling frequency. Evaluation is done at this frequency, and others above mentioned. With RVT, there are no changes when evaluating at different frequencies, while with SSMs rate $r$ is halved for double the inference frequency, etc.
\begin{table*}[t]
\centering
\begin{tabular}{lcccccccc}
\hline
\multirow{2}{*}{Model} & \multirow{2}{*}{Dataset/Size} & \multirow{2}{*}{Strategy} & \multicolumn{5}{c}{Frequency Evaluation (Hz)} & \multirow{2}{*}{Performance Drop} \\
\cline{4-8}
 &  &  & 20 Hz & 40 Hz & 80 Hz & 100 Hz & 200 Hz &  \\
\hline
\multirow{3}{*}{RVT \cite{Gehrig_2023_CVPR}} & Gen1$_{Base}$ & - & 49.52 & 37.16 & 23.25 & 19.36 & 7.83 & 27.62 \\
 & Gen1$_{Small}$ & - & 48.68 & 35.28 & 19.95 & 16.05 & 5.75 & 29.42 \\
 & 1Mpx$_{Base}$ & - & 45.95 & 40.93 & 31.70 & 29.00 & 18.16 & 16.00 \\
\hline
\multirow{6}{*}{S4D \cite{gu_nips2022}} & \multirow{2}{*}{Gen1$_{Base}$} & \( H_2 \) norm & 46.83 & 45.98 & 43.91 & 40.10 & 36.11 & 5.31 \\
 &  & $\alpha = 0.5$ & 47.33 & 46.36 & 44.51 & 40.02 & 35.98 & 5.61 \\
 & \multirow{2}{*}{Gen1$_{Small}$} & \( H_2 \) norm & 45.88 & 45.11 & 41.05 & 38.00 & 34.05 & 6.33 \\
 &  & $\alpha = 0.5$ & 46.30 & 45.21 & 42.11 & 38.61 & 33.00 & 6.57 \\
 & \multirow{2}{*}{1Mpx$_{Base}$} & \( H_2 \) norm & 46.66 & 45.85 & 43.33 & 41.80 & 37.01 & 4.66 \\
 &  & $\alpha = 0.5$ & 47.93 & 46.78 & 44.56 & 41.11 & 36.18 & 5.77 \\
\hline
\multirow{6}{*}{S5 \cite{smith2023simplified}} & \multirow{2}{*}{Gen1$_{Base}$} & \( H_2 \) norm & 48.60 & 47.11 & 46.06 & 43.80 & 40.51 & 4.23 \\
 &  & $\alpha = 0.5$ & 48.48 & 47.34 & 46.11 & 43.23 & 40.03 & 4.30 \\
 & \multirow{2}{*}{Gen1$_{Small}$} & \( H_2 \) norm & 47.33 & 46.32 & 44.03 & 41.12 & 38.98 & 4.72 \\
 &  & $\alpha = 0.5$ & 47.83 & 46.58 & 44.46 & 41.11 & 38.95 & 5.06 \\
 & \multirow{2}{*}{1Mpx$_{Base}$} & \( H_2 \) norm & 48.65 & 47.53 & 47.11 & 46.63 & 40.91 & 3.11 \\
 &  & $\alpha = 0.5$ & 48.35 & 47.60 & 47.21 & 46.50 & 40.80 & 2.82 \\
\hline
\end{tabular}
\caption{Evaluation of RVT, S4D, and S5 across different frequencies on validation datasets}
\label{table:frequencies_comparison}
\end{table*}

Each model in Table~\ref{table:frequencies_comparison} is assessed across datasets and model sizes, including Gen1$_{Base}$, Gen1$_{Small}$, and 1 Mpx$_{Base}$. $Base$ and $Small$ represent two variants of models on the Gen1 and 1 Mpx datasets, the base one being the larger one with 16.5M parameters for the S4D model and 18.2M parameters for the S5 model (as presented in Table~\ref{tab:sota} parameters' column), and small one with 8.8M parameters for S4D and 9.7M parameters for S5 model. The focal point of this analysis is the performance drop, calculated as the average difference between the original performance at 20 Hz and performances at higher frequencies.
A notable aspect of this study is the inherent advantage of the S4D and S5 models due to their incorporation of a learnable timescale parameter. This significantly enhances their adaptability, allowing them to dynamically adjust to varying frequencies. They are further analyzed based on the different operational strategies: the $H_{2}$ norm and bandlimiting with $\alpha = 0.5$. The inclusion of the learnable timescale parameter in these models underscores their capability to maintain performance across a wide range of frequencies.

The analysis reveals that both the $H_{2}$ norm and bandlimiting strategies with $\alpha = 0.5$ offer comparable performance across the assessed frequency ranges. However, a slight edge is observed with the $H_{2}$ norm, particularly at very high frequencies. This marginal superiority can be attributed to the fact that the $H_{2}$ norm approach does not explicitly mask the output's ($C$) matrix columns.

In the Appendix, we study pure-SSM and SSM models in combination with ConvNext \cite{liu2022convnet}.

\label{sec:experiments:ablation:frequencies_comparison}

\vspace{-1ex}

\section{Conclusion}
\label{sec:conclusion}
In this paper, we presented a novel approach for enhancing the adaptability and training efficiency of models designed for event-based vision, particularly in object detection tasks. Our methodology leverages the integration of SSMs with a Vision Transformer (ViT) architecture, creating a hybrid SSM-ViT model. This integration not only addresses the long-standing challenge of performance degradation at varying temporal frequencies but also significantly accelerates the training process.

The key innovation of our work lies in the use of learnable timescale parameters within the SSMs, enabling the model to adapt dynamically to different inference frequencies without necessitating multiple training cycles. This feature represents a substantial advancement over existing methods, which require retraining for different frequencies.

The SSM-ViT model outperforms existing methods by 20 mAP at higher frequencies and exhibits a 33\% increase in training speed.
Furthermore, our introduction of two novel strategies to counteract the aliasing effect (a crucial consideration in high-frequency deployment) further reinforces the model's suitability for real-world applications. These strategies, involving frequency-selective masking and $H_{2}$ norm adjustments, effectively mitigate the adverse effects of aliasing, ensuring the model's reliability across a spectrum of temporal resolutions.
We believe that our approach opens new avenues for research and application in high-speed, dynamic visual environments, setting a new benchmark in the domain of event-based vision.

\section{Acknowledgment}
\vspace{-1.5ex}
\label{sec:acknowledgment}
This work was supported by the European Research Council (ERC) under grant agreement No. 864042 (AGILEFLIGHT).
\clearpage
\setcounter{page}{1}
\maketitlesupplementary

\section*{Contents:}
\begin{itemize}
    \item \textbf{Appendix \ref{suppl:s4d_s5_models}}:  Explanation of S4(D) and S5 models.
    \item \textbf{Appendix \ref{suppl:initializations}}:  Initialization of continuous-time matrices.
    \item \textbf{Appendix \ref{suppl:parscan}}:  Background on Parallel Scans for Linear Recurrences.
    \item \textbf{Appendix \ref{suppl:dsec_dataset}}:  DSEC dataset evaluation.
    \item \textbf{Appendix \ref{suppl:pytorch_impl}}:  PyTorch implementation of Parallel Scan and S5 model.
    \item \textbf{Appendix \ref{suppl:s4_s5_connection}}:  Exploring the S4 and S5 Architectural Link.
    \item \textbf{Appendix \ref{suppl:s5_vit_convnext_pure}}:  SSM-only and SSM-ConvNext \cite{liu2022convnet} models.
    
\end{itemize}

\section{Explanation of S4(D) and S5 models}
\label{suppl:s4d_s5_models}
The Structured State Spaces (S4 and its diagonal variant S4D \cite{gu2022efficiently, gupta2022diagonal}) and Simplified State Space (S5) \cite{smith2023simplified} models are advanced approaches in sequence modeling, each with its unique characteristics.

S4 introduces a sequence model based on the Structured State Space Model (SSM) \cite{gu2020hippo, gu2021combining, gu2023hippo}. It addresses computational bottlenecks in previous work \cite{gu2021combining} by reparameterizing structured state matrices to maintain a hidden state that encodes the long history of input. S4D \cite{gupta2022diagonal} is a variant of the S4 model, simplifying it by using a fully diagonal state matrix. This adaptation preserves the performance of the original model but with a simpler implementation. Gupta \cite{gupta2022diagonal} observed that removing the low-rank part of S4’s HiPPO-LegS matrix \cite{gu2020hippo} results in a diagonal matrix (referred to as the normal-HiPPO matrix) with comparable performance to the original S4. S5 \cite{smith2023simplified} is an evolution of S4, using a multi-input, multi-output (MIMO) single SSM instead of multiple single-input, single-output (SISO) SSMs used in S4 \cite{gu2022efficiently}.

The HiPPO matrix \cite{gu2020hippo}, a non-normal matrix, is decomposed as a sum of a normal and a low-rank matrix. S4 applies new techniques to overcome computational limitations associated with this decomposition. The S5 model simplifies the S4 layer by replacing the bank of many independent SISO SSMs with one MIMO SSM and implementing efficient parallel scans \cite{BlellochTR90}. This shift eliminates the need for convolutional and frequency-domain approaches, making the model purely recurrent and time-domain based.

S4 equates the diagonal matrix case to the computation of a Cauchy kernel, applying to any matrix that can be decomposed as Normal Plus Low-Rank (NPLR). S5 utilizes a diagonal state matrix for efficient computation using parallel scans. It inherits HiPPO initialization schemes from S4, using a diagonal approximation to the HiPPO matrix for comparable performance.

S5 matches the computational complexity of S4 for both online generation and offline recurrence. It handles time-varying SSMs and irregularly sampled observations, which are challenges for the convolutional implementation in S4. S5 has been shown to match or outperform S4 in various long-range sequence modeling tasks, including speech classification and 1-D image classification \cite{smith2023simplified}. This design opens up new possibilities in deep sequence modeling, including handling time-varying SSMs and combining state space layers with attention mechanisms for enhanced performance, which is our contribution along with specific problem and architecture design.

\begin{figure*}[ht]
  \centering
  \includegraphics[width=\textwidth]{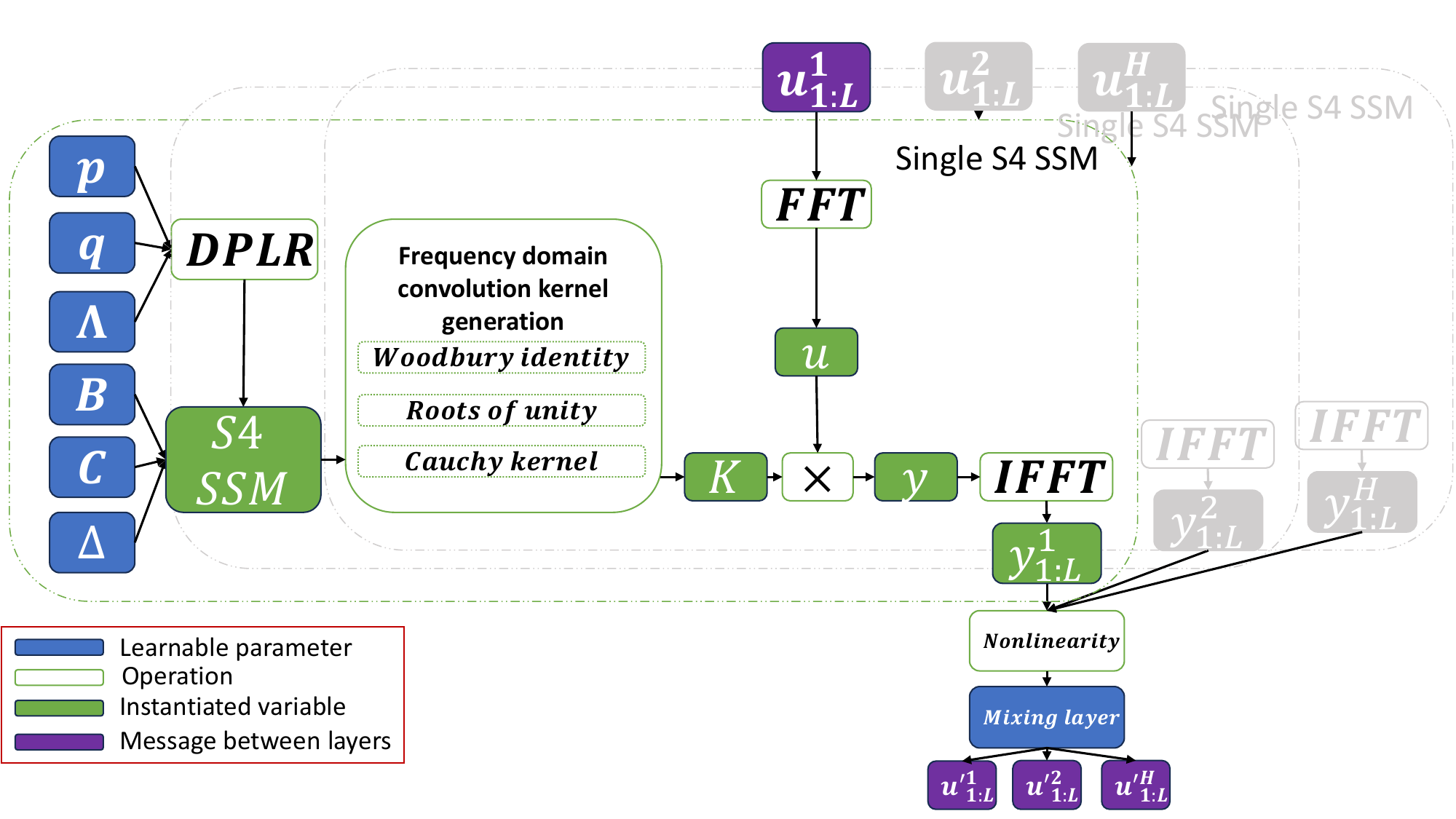}
  \caption{In the S4 layer, each dimension of the input sequence \( u_{1:L} \in \mathbb{R}^{L \times H} \) is processed by a separate SSM. This process involves using a Cauchy kernel to determine the coefficients for frequency domain convolutions. The convolutions, done via FFTs, generate the output \( y_{1:L} \in \mathbb{R}^{L \times H} \) for each SSM. The outputs then go through a nonlinear activation function, which includes a layer that mixes them to produce the final output of the layer.}
  \label{fig:s4}
\end{figure*}

\begin{figure*}[ht]
  \centering
  \includegraphics[width=\textwidth]{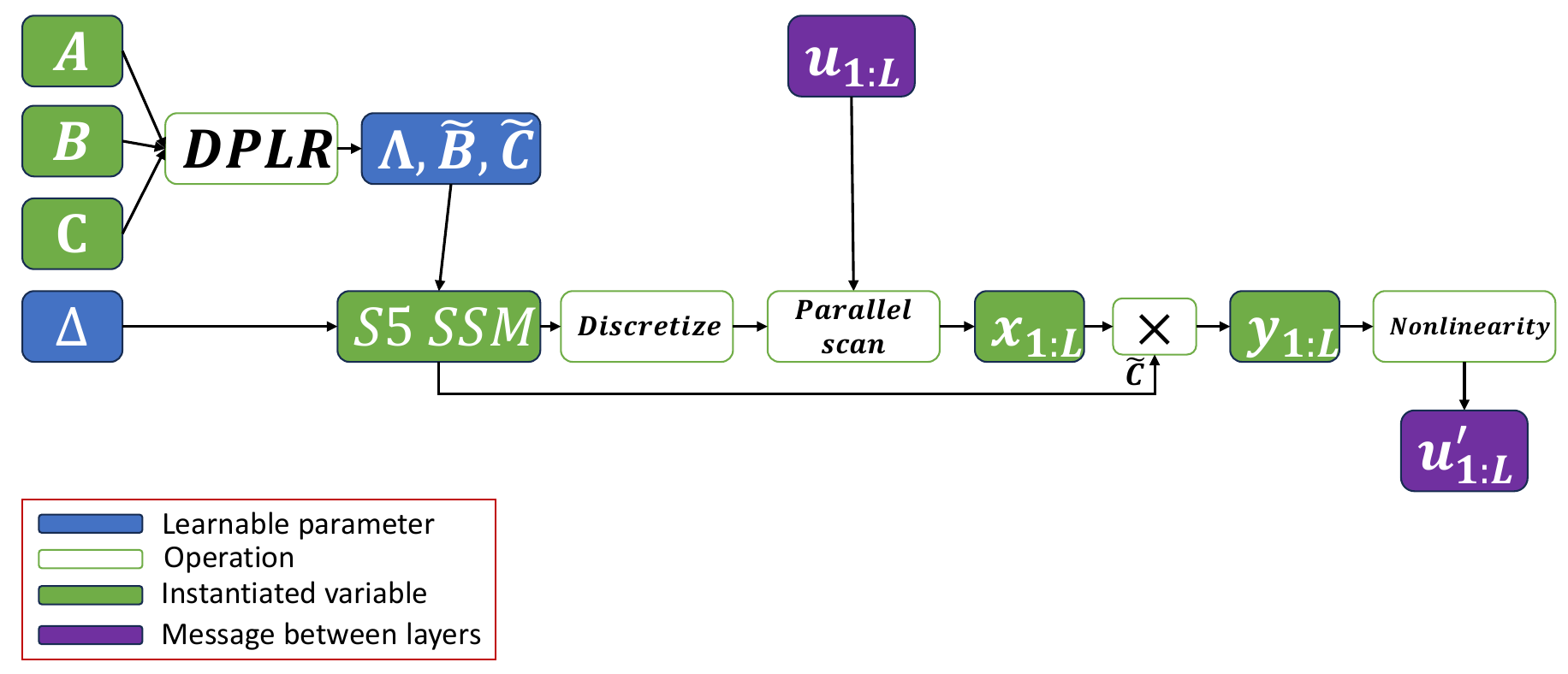}
  \caption{For the S5 layer, a parallel scan technique is employed on a diagonal linear SSM to get the SSM outputs. This approach bypasses the need for frequency domain operations and convolution kernel computations required by S4, resulting in a model that functions in a purely time-domain, recurrent manner. Recurrence is parallelized with the employment of parallel scans \cite{BlellochTR90}.}
  \label{fig:s5}
\end{figure*}

In summary, while S4 introduced a novel reparameterization of state space models for efficient long sequence modeling, S5 builds upon this by simplifying the model structure and computation, leading to a more usable and potentially more flexible approach for various sequence modeling tasks. Both S4 and S5 models can be seen on Figure \ref{fig:s4} and Figure \ref{fig:s5} respectively. 

\section{Initialization of continuous-time matrices}
\label{suppl:initializations}

\subsection{State Matrix Initialization}
\label{subsec:initA}
This subsection elaborates on the initialization process of continuous-time matrices which are crucial since they allow us to discretize them with different time steps to deploy our model at higher inference frequencies. As described by \cite{gu2020hippo}, S4's capacity to handle long-range dependencies stems from employing the HiPPO-LegS matrix, which decomposes the input considering an infinitely long, exponentially diminishing measure \cite{gu2021combining, gu2022efficiently}. The HiPPO-LegS matrix and its corresponding single-input-single-output (SISO) vector are defined as:
\begin{align}
    \left(\mathbf{A_{\mathrm{LegS}}}\right)_{nk} &= - \begin{cases}
                    (2n+1)^{1/2}(2k+1)^{1/2}, &  n>k \\
                    n+1, & n = k \\
                    0, &  n < k
    \end{cases}.\\
    \left(\mathbf{b_{\mathrm{LegS}}}\right)_{n}&=(2n+1)^{\frac{1}{2}}.
\end{align}
Here, the input matrix $\mathbf{B_{\mathrm{LegS}}}\in \mathbb{R}^{N\times H}$ is constructed by concatenating $\mathbf{b_{\mathrm{LegS}}}\in \mathbb{R}^N$ $H$ times.

Theorem 1 of \citet{gu2021combining} establishes that HiPPO matrices from \cite{gu2020hippo}, $\mathbf{A}_{\mathrm{HiPPO}}\in \mathbb{R}^{N \times N}$, can be represented in a normal plus low-rank (NPLR) format, comprising a normal matrix, $\mathbf{A}_{\mathrm{HiPPO}}^{\mathrm{Normal}}=\mathbf{V}\mathbf{\Lambda} \mathbf{V}^* \in \mathbb{R}^{N \times N} $, and a low-rank component:
\begin{align}
    \mathbf{A_{\mathrm{HiPPO}}}  =  \mathbf{A}_{\mathrm{HiPPO}}^{\mathrm{Normal}} - \mathbf{P}\mathbf{Q}^{\top} = \\
    \mathbf{V}\left(\mathbf{\Lambda}- (\mathbf{V}^*\mathbf{P})(\mathbf{V}^*\mathbf{Q})^*\right)\mathbf{V}^* 
\end{align}
with unitary $\mathbf{V}\in \mathbb{C}^{N \times N}$, diagonal $\mathbf{\Lambda}\in \mathbb{C}^{N \times N}$, and low-rank factors $\mathbf{P}, \mathbf{Q}\in \mathbb{R}^{N\times r}$. This equation reveals the conjugation of HiPPO matrices into a diagonal plus low-rank (DPLR) structure. Therefore, the HiPPO-LegS matrix can be expressed through the normal HiPPO-N matrix and the low-rank term $\mathbf{P}_{\mathrm{LegS}}\in \mathbb{R}^N$, as suggested by \citet{goel2022sashimi}:
\begin{align}
    \mathbf{A_{\mathrm{LegS}}}  &= \mathbf{A}_{\mathrm{LegS}}^{\mathrm{Normal}}-\mathbf{P}_{\mathrm{Legs}}\mathbf{P}_{\mathrm{Legs}}^{\top}
\end{align}
where
\begin{align}
    \mathbf{A}_{\mathrm{LegS}_{nk}}^{\mathrm{Normal}} &= 
    -\begin{cases}
    (n+\frac{1}{2})^{1/2}(k+\frac{1}{2})^{1/2}, & n>k\\
    \frac{1}{2}, & n=k\\
    (n+\frac{1}{2})^{1/2}(k+\frac{1}{2})^{1/2}, & n < k
    \end{cases}.\\
    \mathbf{P_{\mathrm{Legs}}}_n &= (n+\frac{1}{2})^{\frac{1}{2}}
\end{align}
We default to initializing the S5 layer state matrix $\mathbf{A}$ as $\mathbf{A}_{\mathrm{LegS}}^{\mathrm{Normal}}\in \mathbb{R}^{P \times P}$ and then decompose it to obtain the initial $\mathbf{\Lambda}$. We often find benefits in initializing $\mathbf{\tilde{B}}$ and $\mathbf{\tilde{C}}$ using $\mathbf{V}$ and its inverse, as detailed below.

Performance improvements on various tasks were noted with the S5 state matrix initialized as block-diagonal \cite{smith2023simplified}, with each diagonal block equaling $\mathbf{A}_{\mathrm{LegS}}^{\mathrm{Normal}}\in \mathbb{R}^{R \times R}$ (where $R < P$). This initialization process also involves decomposing the matrix to obtain $\mathbf{\Lambda}$, as well as $\mathbf{\tilde{B}}$ and $\mathbf{\tilde{C}}$. Even in this case, $\mathbf{\tilde{B}}$ and $\mathbf{\tilde{C}}$ remain densely initialized without constraints to keep $\mathbf{A}$ block-diagonal during learning. The hyperparameter $J$ \cite{smith2023simplified} denotes the number of HiPPO-N blocks on the diagonal for initialization, with $J=1$ indicating the default single HiPPO-N matrix initialization. Further discussion on the rationale behind this block-diagonal approach is in Appendix \ref{app:sec:relation:blocks}.

\subsection{Input, Output, and Feed-through Matrices Initialization}
The input matrix $\mathbf{\tilde{B}}$ and output matrix $\mathbf{\tilde{C}}$ are explicitly initialized using the eigenvectors from the diagonalization of the initial state matrix. We start by sampling $\mathbf{B}$ and $\mathbf{C}$ and then set the complex learnable parameters $\mathbf{\tilde{B}} = \mathbf{V}^{-1}\mathbf{B}$ and $\mathbf{\tilde{C}} = \mathbf{C}\mathbf{V}$.

The feed-through matrix $\mathbf{D}\in \mathbb{R}^H$ is initialized by sampling each element independently from a standard normal distribution.

\subsection{Timescale Initialization}
\label{subsec:initDelta}
Prior research, notably \cite{gupta2022diagonal} and \cite{gu2023hippo}, has underscored the significance of initializing timescale parameters. Explored in detail by \citet{gu2023hippo}, we align with S4 practices and initialize each element of $\log \mathbf{\Delta} \in \mathbb{R}^P$ from a uniform distribution over the interval $[\log \delta_{\min}, \log \delta_{\max})$, with default values set to $\delta_{\min}= 0.001$ and $\delta_{\max}= 0.1$.

\section{Background on Parallel Scans for Linear Recurrences}
\label{suppl:parscan}

This section provides a concise introduction to parallel scans, which are central element for the parallelization of the S5 method implemented in \ref{suppl:pytorch_impl:parallel_scan_implementation}. For an in-depth exploration, refer to \cite{BlellochTR90}.

In essence, a scan operation takes a binary associative operator $\bullet$, adhering to the property $(a \bullet b) \bullet c = a \bullet (b \bullet c)$, and a sequence of $L$ elements $[a_1, a_2, ..., a_{L}]$. The output is the cumulative sequence: $[a_1, \ (a_1 \bullet a_2), \ ..., \ (a_1 \bullet a_2 \bullet ... \bullet a_{L})]$.

The principle behind parallel scans is leveraging the flexible computation order of associative operators. For linear recurrences, parallel scans are applicable on the following state update equations:
\begin{align}
    \mathbf{x}_{k} &= \overline{\mathbf{A}}\mathbf{x}_{k-1} + \overline{\mathbf{B}}\mathbf{u}_{k}, & \mathbf{y}_{k} &= \overline{\mathbf{C}}\mathbf{x}_{k} + \overline{\mathbf{D}}\mathbf{u}_{k} \label{eq:disc_lssr2}
\end{align}
We initiate the process with scan tuples $c_k = (c_{k,a}, c_{k,b}) := (\overline{\mathbf{A}},\enspace \overline{\mathbf{B}}\mathbf{u}_k)$. The binary associative operator, applied to tuples $q_i, q_j$ (initial or intermediate), generates a new tuple  $q_i \bullet q_j := \left(  q_{j,a} \odot q_{i,a}, \enspace q_{j,a} \otimes q_{i,b} + q_{j,b}\right)$, where $\odot$ and $\otimes$ represent matrix-matrix and matrix-vector multiplication, respectively. With adequate processing power, this parallel scan approach can compute the linear recurrence of (\ref{eq:disc_lssr2}) in $O(\log L)$ sequential steps, thereby significantly enhancing computational efficiency~\cite{BlellochTR90}.


\section{DSEC dataset evaluation}
\label{suppl:dsec_dataset}
To test the generalizability of our model, we perform inference on the DSEC dataset \cite{Gehrig2021ral_a} with a model trained on 1 Mpx dataset \cite{perot_nips2020}. Some results are shown below:
\begin{figure}[H]
  \centering
  \begin{subfigure}[b]{0.47\columnwidth} 
    \includegraphics[width=\linewidth]{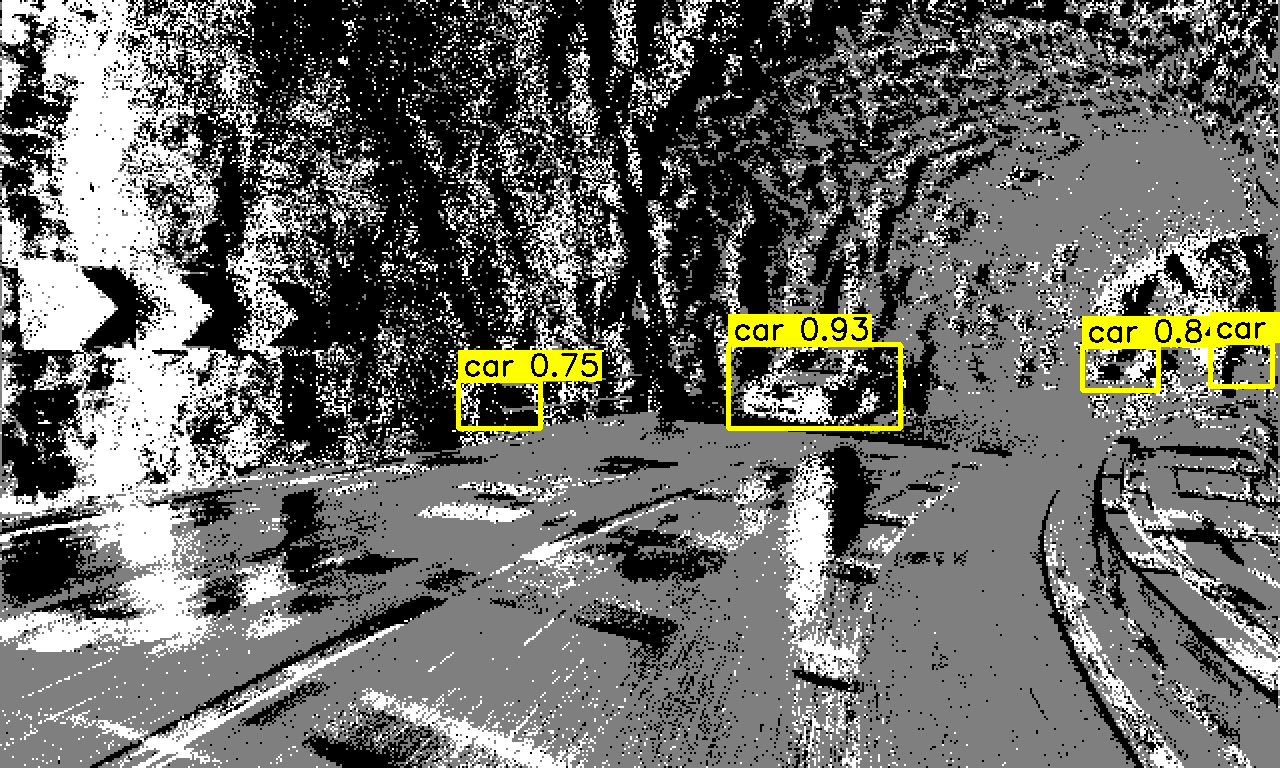}
    \caption{$interlaken\_00\_b$}
    \label{fig:sub1}
  \end{subfigure}
  \hspace{2mm} 
  \begin{subfigure}[b]{0.47\columnwidth} 
    \includegraphics[width=\linewidth]{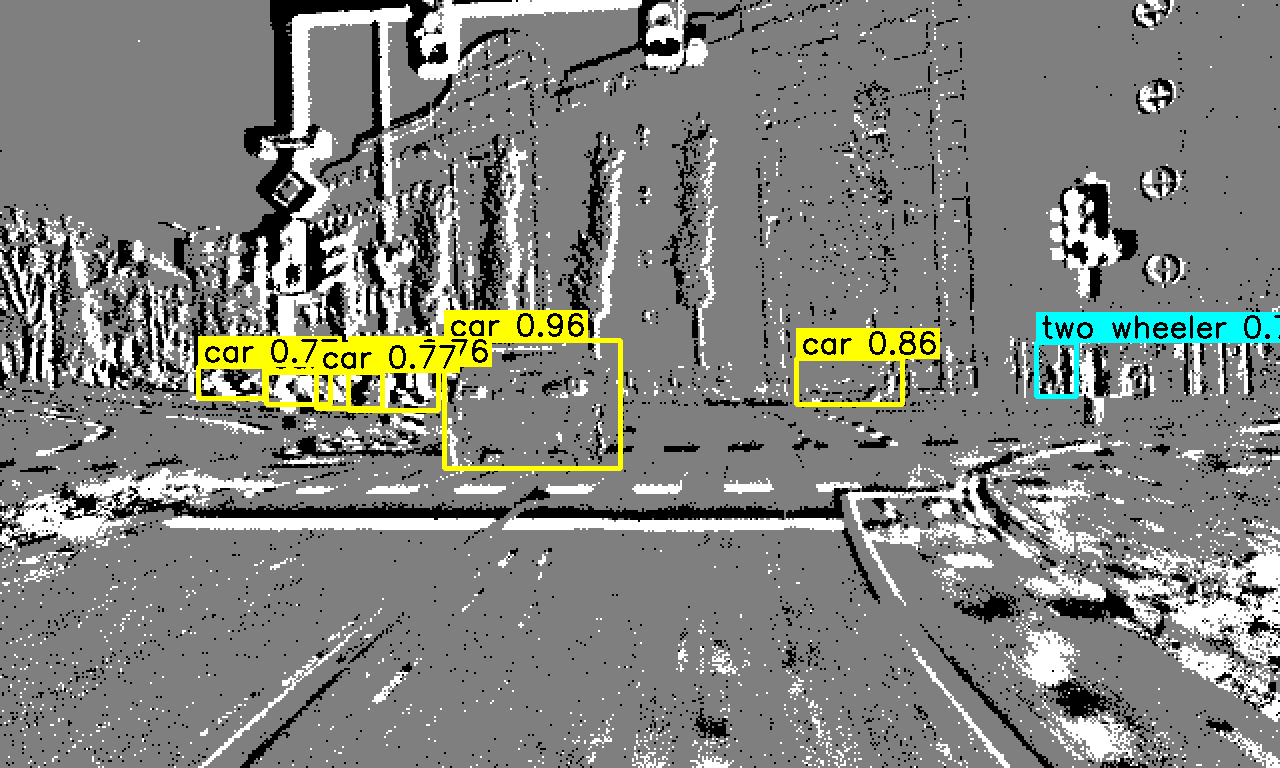}
    \caption{$zurich\_city\_00\_a$}
    \label{fig:sub2}
  \end{subfigure}
  
  \vspace{3mm}
  
  \begin{subfigure}[b]{0.47\columnwidth} 
    \includegraphics[width=\linewidth]{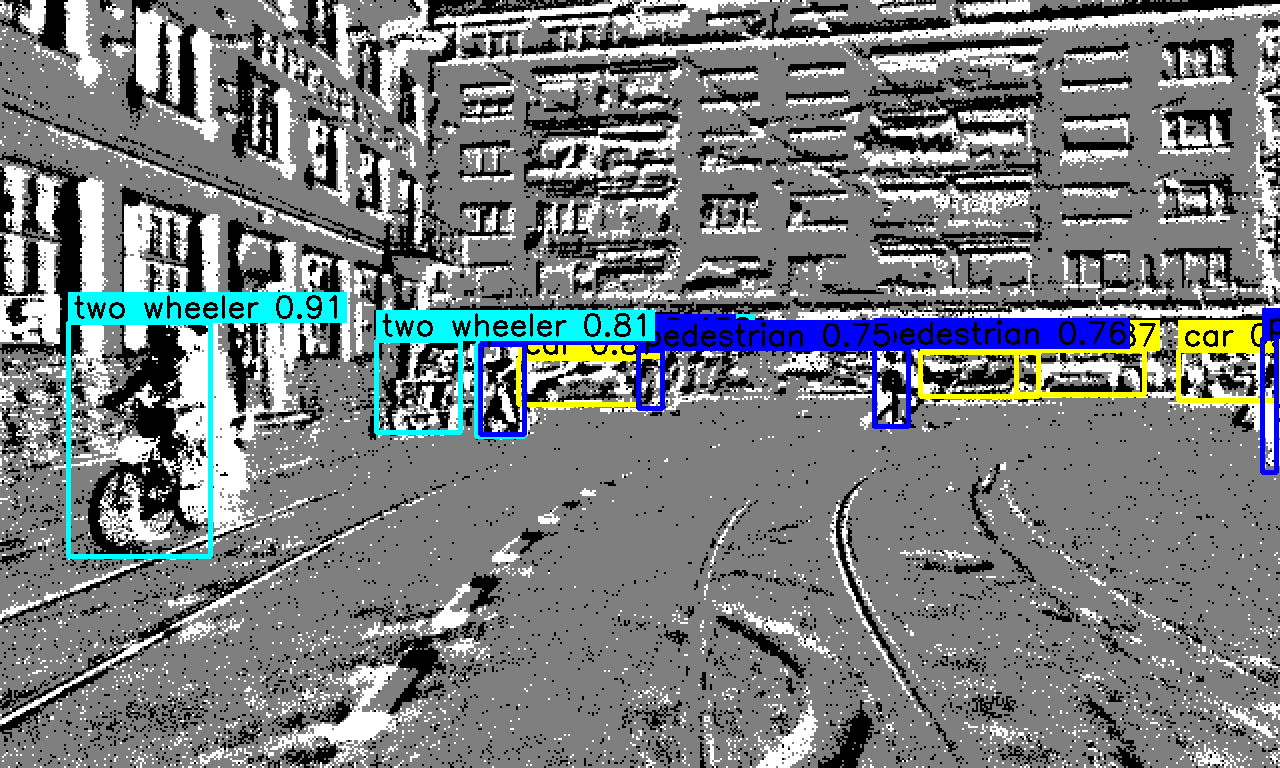}
    \caption{$zurich\_city\_00\_b$}
    \label{fig:sub3}
  \end{subfigure}
  \hspace{2mm} 
  \begin{subfigure}[b]{0.47\columnwidth} 
    \includegraphics[width=\linewidth]{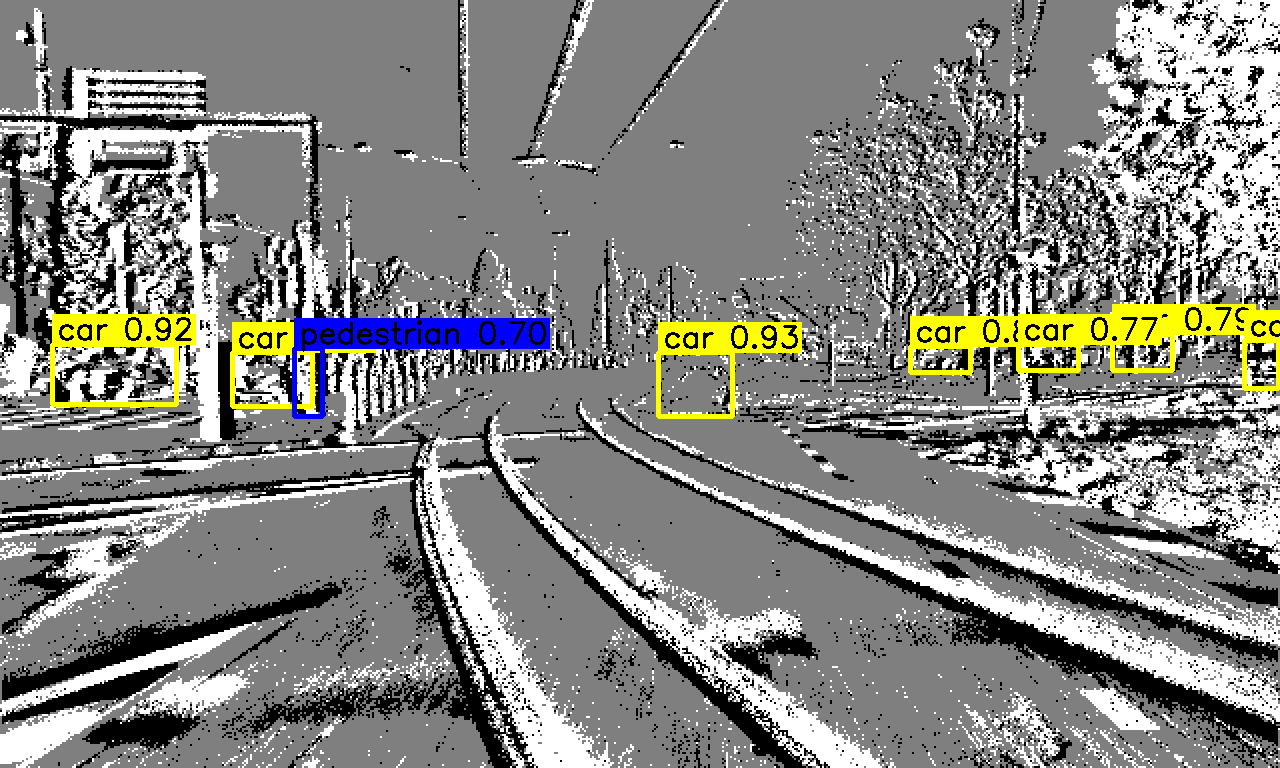}
    \caption{$zurich\_city\_07\_a$}
    \label{fig:sub4}
  \end{subfigure}
  \caption{Detections on DSEC dataset \cite{Gehrig2021ral_a} with model trained on 1 Mpx dataset \cite{perot_nips2020}. Names of the specific DSEC scenes are in the subcaptions.}
  \label{fig:2x2matrix}
\end{figure}

\section{PyTorch implementation of Parallel Scan and S5 model}
\label{suppl:pytorch_impl}
In the following subsections, we give implementations of parallel scans and S5 layer in PyTorch.
\subsection{Parallel Scan Operation}
\label{suppl:pytorch_impl:parallel_scan_implementation}
\begin{minipage}{13.8cm}
\begin{center}
\begin{lstlisting}[language=Python, basicstyle=\scriptsize\ttfamily, showlines=true, caption=Implementation of efficient parallel scan used in S5 model.]
from typing import Callable
import torch
from torch.utils._pytree import tree_flatten, tree_unflatten
from functools import partial

def safe_map(f, *args):
    args = list(map(list, args))
    n = len(args[0])
    return list(map(f, *args))
    
def combine(tree, operator, a_flat, b_flat):
    a = tree_unflatten(a_flat, tree)
    b = tree_unflatten(b_flat, tree)
    c = operator(a, b)
    c_flat, _ = tree_flatten(c)
    return c_flat

def _scan(tree, operator, elems, axis: int):
    num_elems = elems[0].shape[axis]

    if num_elems < 2:
        return elems

    reduced_elems = combine(tree, operator,
        [torch.ops.aten.slice(elem, axis, 0, -1, 2) for elem in elems],
        [torch.ops.aten.slice(elem, axis, 1, None, 2) for elem in elems],
    )
    
    # Recursively compute scan for partially reduced tensors.
    odd_elems = _scan(tree, operator, reduced_elems, axis)

    if num_elems % 2 == 0:
        even_elems = combine(tree, operator,
            [torch.ops.aten.slice(e, axis, 0, -1) for e in odd_elems],
            [torch.ops.aten.slice(e, axis, 2, None, 2) for e in elems],)
    else:
        even_elems = combine(tree, operator, odd_elems,
            [torch.ops.aten.slice(e, axis, 2, None, 2) for e in elems],)

    # The first element of a scan is the same as the first element of the original `elems`.
    even_elems = [
        torch.cat([torch.ops.aten.slice(elem, axis, 0, 1), result], dim=axis)
        if result.shape.numel() > 0 and elem.shape[axis] > 0
        else result
        if result.shape.numel() > 0
        else torch.ops.aten.slice(
            elem, axis, 0, 1
        )  # Jax allows/ignores concat with 0-dim, Pytorch does not
        for (elem, result) in zip(elems, even_elems)
    ]
    return list(safe_map(partial(_interleave, axis=axis), even_elems, odd_elems))

def associative_scan(operator: Callable, elems, axis: int = 0, reverse: bool = False):
    elems_flat, tree = tree_flatten(elems)

    if reverse:
        elems_flat = [torch.flip(elem, [axis]) for elem in elems_flat]
    num_elems = int(elems_flat[0].shape[axis])
    scans = _scan(tree, operator, elems_flat, axis)

    if reverse:
        scans = [torch.flip(scanned, [axis]) for scanned in scans]

    return tree_unflatten(scans, tree)

\end{lstlisting}
\end{center}
\end{minipage}

\clearpage

\subsection{S5 model}
\label{suppl:pytorch_impl:s5_implementation}

\begin{minipage}{13.8cm}
\begin{center}
\begin{lstlisting}[language=Python, basicstyle=\scriptsize\ttfamily, showlines=true, caption=PyTorch implementation to apply a single S5 layer to a batch of input sequences.]
import torch
from typing import Tuple

def discretize_bilinear(Lambda, B_tilde, Delta):
    """ Discretize a diagonalized, continuous-time linear SSM
        using bilinear transform method.
        Args:
            Lambda (float32): diagonal state matrix              (P, 2)
            B_tilde (complex64): input matrix                    (P, H)
            Delta (float32): discretization step sizes           (P,)
        Returns:
            discretized Lambda_bar (float32), B_bar (float32)  (P, 2), (P, H, 2)
    """
    Lambda = torch.view_as_complex(Lambda)

    Identity = torch.ones(Lambda.shape[0], device=Lambda.device)
    BL = 1 / (Identity - (Delta / 2.0) * Lambda)
    Lambda_bar = BL * (Identity + (Delta / 2.0) * Lambda)
    B_bar = (BL * Delta)[..., None] * B_tilde

    Lambda_bar = torch.view_as_real(Lambda_bar)
    B_bar = torch.view_as_real(B_bar)

    return Lambda_bar, B_bar

@torch.jit.script
def binary_operator(
    q_i: Tuple[torch.Tensor, torch.Tensor], q_j: Tuple[torch.Tensor, torch.Tensor]
):
    """Binary operator for parallel scan of linear recurrence. Assumes a diagonal matrix A.
    Args:
        q_i: tuple containing A_i and Bu_i at position i       (P,), (P,)
        q_j: tuple containing A_j and Bu_j at position j       (P,), (P,)
    Returns:
        new element ( A_out, Bu_out )
    """
    A_i, b_i = q_i
    A_j, b_j = q_j

    return A_j * A_i, torch.addcmul(b_j, A_j, b_i)

def apply_ssm(
    Lambda_bars: torch.Tensor, B_bars, C_tilde, D, input_sequence, prev_state, bidir: bool = False
):
    """ Compute the LxH output of discretized SSM given an LxH input.
        Args:
            Lambda_bars (float32): discretized diagonal state matrix    (P, 2)
            B_bars      (float32): discretized input matrix             (P, H, 2)
            C_tilde    (float32): output matrix                         (H, P, 2)
            input_sequence (float32): input sequence of features        (L, H)
            prev_state (complex64): hidden state                        (H,)
        Returns:
            ys (float32): the SSM outputs (S5 layer preactivations)     (L, H)
            xs (complex64): hidden state                                (H,)
    """
    B_bars, C_tilde, Lambda_bars = as_complex(B_bars), as_complex(C_tilde), as_complex(Lambda_bars)

    cinput_sequence = input_sequence.type(Lambda_bars.dtype)  # Cast to correct complex type
    Bu_elements = torch.vmap(lambda u: B_bars @ u)(cinput_sequence)

    if Lambda_bars.ndim == 1:  # Repeat for associative_scan
        Lambda_bars = Lambda_bars.tile(input_sequence.shape[0], 1)
    
    Lambda_bars[0] = Lambda_bars[0] * prev_state
    _, xs = associative_scan(binary_operator, (Lambda_bars, Bu_elements))

    if bidir:
        _, xs2 = associative_scan(
            binary_operator, (Lambda_bars, Bu_elements), reverse=True
        )
        xs = torch.cat((xs, xs2), axis=-1)

    Du = torch.vmap(lambda u: D * u)(input_sequence)
    return torch.vmap(lambda x: (C_tilde @ x).real)(xs) + Du, xs[-1]
\end{lstlisting}
\end{center}
\end{minipage}

\clearpage

\begin{minipage}{13.8cm}
\begin{center}
\begin{lstlisting}[language=Python, basicstyle=\scriptsize\ttfamily, showlines=true, caption=Raw S5 operator that can be instantiated and used as block in any architecture.]
import torch

class S5(torch.nn.Module):
    def __init__(...):
        self.seq = S5SSM(...)
        ...
        
    def forward(self, signal, prev_state, step_scale: float | torch.Tensor = 1.0):
        if not torch.is_tensor(step_scale):
            # Duplicate across batchdim
            step_scale = torch.ones(signal.shape[0], device=signal.device) * step_scale

        return torch.vmap(lambda s, ps, ss: self.seq(s, prev_state=ps, step_scale=ss))(signal, prev_state, step_scale)

class S5SSM(torch.nn.Module):
    def __init__(...):
        self.degree = degree
        self.discretize = self.discretize_bilinear
        ...

    def get_BC_tilde(self):
        match self.bcInit:
            case "dense_columns" | "dense" | "complex_normal":
                B_tilde = as_complex(self.B)
                C_tilde = self.C
            case "factorized":
                B_tilde = self.BP @ self.BH.T
                C_tilde = self.CH.T @ self.CP
        return B_tilde, C_tilde
        
    def forward(self, signal, prev_state, step_scale: float | torch.Tensor = 1.0):
        B_tilde, C_tilde = self.get_BC_tilde()
        if self.degree != 1:
            assert (
                B_bar.shape[-2] == B_bar.shape[-1]
            ), "higher-order input operators must be full-rank"
            B_bar **= self.degree

        step = step_scale * torch.exp(self.log_step)
        
        Lambda_bars, B_bars = self.discretize(self.Lambda, B_tilde, step)
        
        return apply_ssm(Lambda_bars, B_bars, C_tilde, self.D, signal, prev_state, bidir=self.bidir)


\end{lstlisting}
\end{center}
\end{minipage}

\clearpage

\section{Exploring the S4 and S5 Architectural Link}
\label{suppl:s4_s5_connection}
This section delves into the relationship between the S4 and S5 architectures, which is instrumental in the evolution of more efficient architectures and the expansion of theoretical insights from preceding research \cite{gu2022efficiently, smith2023simplified, gu2020hippo}.

Our analysis is segmented into three distinct parts:
\begin{enumerate}
\item We utilize the linear nature of these systems to explain that the latent states generated by the S5 SSM are effectively a linear combination of those produced by the $H$ SISO S4 SSMs. Moreover, the outputs of the S5 SSM represent an additional linear transformation of these states (\ref{app:sec:relation:equiv}).
\item For the SISO scenario, as $N$ becomes significantly large, the dynamics derived from a (non-diagonalizable) HiPPO-LegS matrix can be accurately approximated by the (diagonalizable) normal component of the HiPPO-LegS matrix. This is expanded to encompass the MIMO context, providing a rationale for initializing with the HiPPO-N matrix and consequently enabling efficient parallel scans (\ref{app:sec:relation:blocks}).
\item We conclude with a discussion that S5, through strategic initialization of its state matrix, can emulate multiple independent S4 systems, thus easing previously held assumptions (\ref{app:sec:relation:blocks}). We also discuss the set of timescale parameters, which have been observed to enhance performance (\ref{app:sec:relation:timescales}).
\end{enumerate}
It is important to note that many of these results are derived directly from the linearity of the recurrence.

\subsection{Underlying Assumptions}
\label{app:sec:relation:assumptions}
The forthcoming sections are predicated on the following assumptions unless stated otherwise:

\begin{itemize}
    \item We exclusively consider sequence maps that are $H$-dimensional to $H$-dimensional.
    \item The state matrix for each S4 SSM is the same, denoted as $\mathbf{A}^{(h)} = \mathbf{A} \in \mathbb{C}^{N \times N}$.
    \item We assert that the timescales for each S4 SSM are consistent, represented by $\Delta^{(h)} = \Delta \in \mathbb{R}_+$.
    \item S5 employs the identical state matrix $\mathbf{A}$ as in S4 \cite{gu2022efficiently}, implying that the S5's latent size $P$ is such that $P=N$. Additionally, it is presumed that the S5 input matrix is a horizontal concatenation of the S4's column input vectors, expressed as $\mathbf{B} \triangleq \left[ \mathbf{B}^{(1)} \mid \ldots \mid \mathbf{B}^{(H)} \right]$.
\end{itemize}

\subsection{Distinct Output Projections from Equivalent Dynamics}
\label{app:sec:relation:equiv}
\label{ass:prop:equiv}
In an S5 layer characterized by state matrix $\mathbf{A}$, input matrix $\mathbf{B}$, and an output matrix $\mathbf{C}$, and an S4 layer, where each of the $H$ individual S4 SSMs possesses a state matrix $\mathbf{A}$ and input vector $\mathbf{B}^{(h)}$, the outputs of the S5 SSM, $\mathbf{y}_k$, are equivalent to a linear combination of the latent states from the $H$ S4 SSMs, $\mathbf{y}_k = \mathbf{C}^{\mathrm{equiv}}\mathbf{x}_k^{(1:H)}$, where $\mathbf{C}^{\mathrm{equiv}} = \left[\ \mathbf{C}\ \cdots\ \mathbf{C}\ \right]$, provided that both layers are discretized using identical timescales.

\begin{proof}
Considering a singular S4 SSM, the discretized latent states in relation to the input sequence $\mathbf{u}_{1:L} \in \mathbb{R}^{L \times H}$ are described as:
\begin{equation}
    \mathbf{x}_{k}^{(h)} = \sum\nolimits_{i=1}^{k} \overline{\mathbf{A}}^{k-i} \overline{\mathbf{B}}^{(h)} u_i^{(h)}.  \label{equ:relation:s4}
\end{equation}
For the S5 layer, the latent states are given by:
\begin{equation}
    \mathbf{x}_{k} = \sum\nolimits_{i=1}^{k} \overline{\mathbf{A}}^{k-i} \overline{\mathbf{B}} \mathbf{u}_i,  \label{equ:relation:s5}
\end{equation}
where $\overline{\mathbf{B}}$ is defined as $\overline{\mathbf{B}} \triangleq \left[ \overline{\mathbf{B}}^{(1)} \mid \ldots \mid \overline{\mathbf{B}}^{(H)} \right]$ and $\mathbf{u}_i$ is $\left[ u^{(1)}_i , \ldots , u^{(H)}_i \right]^{\top}$.

Observation leads to:
\begin{equation} 
    \mathbf{x}_{k} = \sum\nolimits_{h=1}^H \mathbf{x}_{k}^{(h)},  \label{app:equ:relation:equiv_dynamics}
\end{equation}
This outcome directly stems from the linearity of Equations \eqref{equ:relation:s4} and \eqref{equ:relation:s5}, indicating that the MIMO S5 SSM states are equivalent to the sum of the states from the $H$ SISO S4 SSMs.

The output matrix $\mathbf{C}$ for S5 is a singular dense matrix:
\begin{equation}
\mathbf{y}_k = \mathbf{C} \mathbf{x}_k. \label{app:equ:relation:_s5_y}
\end{equation}
Substituting the relationship from \eqref{app:equ:relation:equiv_dynamics} into \eqref{app:equ:relation:_s5_y} enables expressing the outputs of the MIMO S5 SSM in terms of the states of the $H$ SISO S4 SSMs:
\begin{align}
    \mathbf{y}_k &= \mathbf{C} \sum\nolimits_{h=1}^H \mathbf{x}^{(h)}_k = \sum\nolimits_{h=1}^H \mathbf{C} \mathbf{x}^{(h)}_k  \label{equ:relation:s5_y}.
\end{align}
Defining the vertical concatenation of the $H$ S4 SSM state vectors as $\mathbf{x}_k^{(1:H)} = \left[ \mathbf{x}_k^{(1)^{\top}} , \ldots , \mathbf{x}_k^{(H)^{\top}} \right]^{\top}$, we ascertain that the S5 SSM outputs can be written as:
\begin{equation}
\mathbf{y}_k = \mathbf{C}^{\mathrm{equiv}}\mathbf{x}_k^{(1:H)}, \quad \mathrm{with} \quad \mathbf{C}^{\mathrm{equiv}} = \left[\ \mathbf{C} \mid \cdots \mid \mathbf{C}\ \right], \label{app:equ:s5:c-equiv}
\end{equation}
thereby confirming their equivalence to a linear combination of the $HN$ states computed by the $H$ S4 SSMs.
\end{proof}

This proof demonstrates that the outputs of the S5 SSM, under the specified constraints, can be interpreted as a linear combination of the latent states generated by $H$ similarly constrained S4 SSMs, sharing identical state matrices and timescale parameters. However, it does not imply that the outputs of the S5 SSM are directly identical to those of the effective block-diagonal S4 SSM; indeed, they differ, as will be further clarified in the analysis of the S4 layer. 

Assuming the output vector for each S4 SSM corresponds to a row in the S5 output matrix, i.e., $\mathbf{C} = \left[ \mathbf{C}^{(1)^{\top}} \mid \ldots \mid \mathbf{C}^{(H)^{\top}} \right]^{\top}$, the output of each S4 SSM can be expressed as:
\begin{align}
y_k^{(h)} = \mathbf{C}^{(h)} \mathbf{x}_k^{(h)},
\end{align}
where $y_k^{(h)} \in \mathbb{R}$. The effective output matrix operating on the entire latent space in S4 is:
\begin{equation}
y_k^{(h)} = \left( \mathbf{C}^{\mathrm{S4}} \mathbf{x}_k \right)^{(h)} \label{app:equ:relation:s4_y}
\end{equation}
By comparing \eqref{app:equ:s5:c-equiv} and \eqref{app:equ:relation:s4_y}, the distinction in the equivalent output matrices employed by both layers becomes clear:
\begin{align}
    \mathbf{C}^{\mathrm{S4}} &= \left[ \begin{array}{ccc} 
        \mathbf{C}^{(1)} & \cdots & \mathbf{0} \\ 
        \vdots & \ddots & \vdots \\ 
        \mathbf{0} & \cdots & \mathbf{C}^{(H)}  
    \end{array} \right], \\
    \mathbf{C}^{\mathrm{equiv}} &= \left[ \begin{array}{ccc} 
        \mathbf{C}^{(1)} & \cdots & \mathbf{C}^{(1)} \\ 
        \vdots & \ddots & \vdots \\ 
        \mathbf{C}^{(H)} & \cdots & \mathbf{C}^{(H)}  
    \end{array} \right] = \left[ \mathbf{C} \mid \cdots \mid \mathbf{C} \right]. \label{equ:relation:c_eff}
\end{align}

In S4, the effective output matrix comprises independent vectors on the diagonal, while in S5, it uniformly connects dense output matrices across the $H$ S4 SSMs. Thus, S5 can be seen as defining a different projection of the $H$ independent SISO SSMs than S4 does. Both projection matrices possess an identical parameter count.

Despite variations in projection, the interpretability of the latent dynamics in S5 as a linear projection from S4's latent dynamics suggests a promising approach. This observation leads to the hypothesis that initializing the state dynamics in S5 with the HiPPO-LegS matrix, the same as in the method employed in S4 \cite{gu2022efficiently}, may yield similarly effective results.

It remains an open question whether one method consistently surpasses the other in terms of expressiveness. It's also important to underscore that practical implementation of S4 and S5 would not directly utilize the block diagonal matrix and repeated matrix, respectively, as described in Equation \ref{equ:relation:c_eff}. These matrices serve primarily as theoretical tools to explain the conceptual equivalence between S4 and S5 models.

\subsection{Relaxing the Assumptions}
\label{app:sec:relation:blocks}
Consider an instance where the S5 SSM state matrix is configured in a block-diagonal form. In such a scenario, an S5 SSM with a latent size of $JN=\mathcal{O}(H)$ would employ a block-diagonal matrix $\mathbf{A} \in \mathbb{R}^{JN \times JN}$, complemented by dense matrices $\mathbf{B} \in \mathbb{R}^{JN \times H}$ and $\mathbf{C} \in \mathbb{R}^{H \times JN}$, and $J$ distinct timescale parameters $\mathbf{\Delta} \in \mathbb{R}^{J}$. The latent state $\mathbf{x}_k \in \mathbb{R}^{JN}$ of this system can be divided into $J$ distinct states $\mathbf{x}_k^{(j)} \in \mathbb{R}^N$. Consequently, this allows the decomposition of the system into $J$ individual subsystems, with each subsystem discretized using a respective $\Delta^{(j)}$. The discretization can be represented as:
\begin{equation}
\overline{\mathbf{A}} = \left[ \begin{array}{ccc} \overline{\mathbf{A}}^{(1)} &   &   \\   & \ddots &   \\   &   & \overline{\mathbf{A}}^{(J)}  \end{array} \right], \quad \overline{\mathbf{B}} = \left[ \begin{array}{ccc} \overline{\mathbf{B}}^{(1)} \\ \vdots \\ \overline{\mathbf{B}}^{(J)}  \end{array} \right],
\end{equation}
\begin{equation}
\quad \mathbf{C} = \left[ \begin{array}{ccc} \mathbf{C}^{(1)} \mid  \cdots \mid \mathbf{C}^{(J)}  \end{array} \right],
\end{equation}
where $\overline{\mathbf{A}}^{(j)} \in \mathbb{R}^{N \times N}$, $\overline{\mathbf{B}}^{(j)} \in \mathbb{R}^{N \times H}$, and $\mathbf{C}^{(j)} \in \mathbb{R}^{H \times N}$. This division implies that the system can be interpreted as $J$ independent $N$-dimensional S5 SSM subsystems, with the total output being the sum of the outputs from these $J$ subsystems:
\begin{align}
    \mathbf{y}_k &= \mathbf{C}\mathbf{x}_k = \sum_{j=1}^J \mathbf{C}^{(j)}\mathbf{x}_k^{(j)}.
\end{align}
The dynamics of each of these $J$ S5 SSM subsystems can be correlated to the dynamics of a distinct S4 system. Each of these S4 systems possesses its unique set of tied S4 SSMs, including separate state matrices, timescale parameters, and output matrices. Hence, the outputs of a $JN$-dimensional S5 SSM effectively correspond to the linear combination of the latent states from $J$ different S4 systems. This realization opens the possibility of initializing a block-diagonal S5 state matrix with multiple HiPPO-N matrices across its blocks, rather than a singular, larger HiPPO-N matrix.

\subsection{Timescale Parameterization}
\label{app:sec:relation:timescales}
This section delves into the parameterization nuances of the timescale parameters $\mathbf{\Delta}$. S4 possesses the capability to learn distinct timescale parameters for each S4 SSM \cite{gu2022efficiently}, thereby accommodating various data timescales. Additionally, the initial setting of these timescales is crucial, as highlighted in the works of \cite{gu_nips2022} and \cite{gupta2022diagonal}. Relying solely on a single initial parameter might result in suboptimal initialization. The previous discussion in this paper proposes the learning of $J$ separate timescale parameters, corresponding to each of the $J$ subsystems. However, empirical evidence indicates superior performance when employing $P$ distinct timescale parameters \cite{smith2023simplified, gu2022efficiently}, one assigned to each state.

This approach can be interpreted in two ways. Firstly, it may be seen as assigning a unique scaling factor to each eigenvalue within the diagonalized system framework. Alternatively, it could be considered a strategy to increase the diversity of timescale parameters at the initialization phase, thereby mitigating the risks associated with inadequate initialization. It is noteworthy that the system has the potential to \emph{learn} to operate with a singular timescale \cite{smith2023simplified}, achieved by equalizing all timescale values.

\section{SSM-only and SSM-ConvNext models}
\label{suppl:s5_vit_convnext_pure}
To study the performance of SSM-only models, we replaced the attention block with a 2D-SSM block and trained it on Gen1 and 1 Mpx datasets. We obtain 46.10 \textit{mAP} and 45.74 \textit{mAP} which is lower than the original S5-ViT-B model's performance, as it can be seen in Table~\ref{tab:rebuttal_map_scores}. To evaluate a CNN-based backbone, we replace the attention block with the ConvNext block \cite{liu2022convnet} showing that ViT achieves better performance than the CNN structure.
\begin{table}[htbp]
\centering
\begin{tabular}{lcc}
\hline
\textbf{Model} & \textbf{mAP$_{Gen1}$} & \textbf{mAP$_{1 Mpx}$} \\
\hline
S5-ViT-B & 47.71 & 47.80 \\
S5-ConvNext-B & 45.92 & 45.66 \\
S5-SSM2D-B & 46.10 & 45.74 \\
\hline
\end{tabular}
\caption{\textbf{Comparison of mAP scores for Gen1 and 1 Mpx datasets across different base models.}}
\label{tab:rebuttal_map_scores}
\end{table}


{
    \small
    \bibliographystyle{ieeenat_fullname}
    \bibliography{main}
}


\end{document}